\documentclass{article}
\PassOptionsToPackage{table}{xcolor}
\usepackage{xcolor}

\usepackage{microtype}
\usepackage{graphicx}
\usepackage{subcaption}
\usepackage{booktabs} 

\usepackage{hyperref}

\usepackage[accepted]{icml2025}

\usepackage{amsmath}
\usepackage{amssymb}
\usepackage{mathtools}
\usepackage{amsthm}

\usepackage[capitalize,noabbrev]{cleveref}

\usepackage{graphicx}
\usepackage{enumitem}
\usepackage{caption}
\usepackage{subcaption}
\usepackage{listings}
\usepackage{courier}
\usepackage{enumitem}

\theoremstyle{plain}
\newtheorem{theorem}{Theorem}[section]

\newtheorem{lemma}[theorem]{Lemma}

\theoremstyle{definition}

\theoremstyle{remark}

\usepackage[textsize=tiny]{todonotes}

\usepackage{array}
\newcommand{\PreserveBackslash}[1]{\let\temp=\\#1\let\\=\temp}
\newcolumntype{C}[1]{>{\PreserveBackslash\centering}p{#1}}
\newcolumntype{L}[1]{>{\PreserveBackslash\raggedright}p{#1}}

\usepackage{etoolbox}
\usepackage{pgf} 
\definecolor{high}{HTML}{76f013}  
\definecolor{low}{HTML}{ec462e}  

\newcommand{\gradientcell}[6]{
    \ifdimcomp{#1pt}{>}{#3 pt}{#1}{
        \ifdimcomp{#1pt}{<}{#2 pt}{#1}{
            \pgfmathparse{int(round(100*(#1/(#3-#2))-(#2*(100/(#3-#2)))))}
            \xdef\tempa{\pgfmathresult}
            \cellcolor{#5!\tempa!#4!#6} #1
    }}
}

\icmltitlerunning{Policy Filtration for RLHF to Mitigate Noise in Reward Models}

\begin{document}

\twocolumn[
\icmltitle{Policy Filtration for RLHF to Mitigate Noise in Reward Models}



\icmlsetsymbol{equal}{*}

\begin{icmlauthorlist}
\icmlauthor{Chuheng Zhang}{equal,msra}
\icmlauthor{Wei Shen}{equal,ind}
\icmlauthor{Li Zhao}{msra}
\icmlauthor{Xuyun Zhang}{mac}
\icmlauthor{Xiaolong Xu}{nust}
\icmlauthor{Wanchun Dou}{nju}
\icmlauthor{Jiang Bian}{msra}
\end{icmlauthorlist}

\icmlaffiliation{msra}{Microsoft Research}
\icmlaffiliation{ind}{Independent Researcher}
\icmlaffiliation{mac}{Macquarie University}
\icmlaffiliation{nust}{Nanjing University of Information Science and Technology}
\icmlaffiliation{nju}{State Key Laboratory for Novel Software Technology, Nanjing University}

\icmlcorrespondingauthor{Chuheng Zhang}{zhangchuheng123@live.com}
\icmlcorrespondingauthor{Wei Shen}{shenwei0917@126.com}

\icmlkeywords{AI, LLM, RLHF}

\vskip 0.3in
]



\printAffiliationsAndNotice{\icmlEqualContribution} 

\begin{abstract}
While direct policy optimization methods exist, pioneering LLMs are fine-tuned with reinforcement learning from human feedback (RLHF) to generate better responses under the supervision of a reward model learned from preference data. One major challenge of RLHF is the inaccuracy of the intermediate reward model, especially in the tasks that requires complex reasoning for the reward model to score a response. We find that the reliability of the reward model varies across responses assigned with different rewards. This motivates us to filter the samples whose rewards may be unreliable to improve the signal-to-noise ratio during policy learning, resulting in Policy Filtration for Proximal Policy Optimization (PF-PPO). To choose a proper policy filtering strategy, we use the coefficient of determination ($R^2$) between the rewards and actual scores on filtered samples as the metrics to help us find promising strategies since it measures how well the rewards filtered by PF-PPO indicate real performance. We provide extensive experiments to validate the effectiveness of PF-PPO in code generation and math reasoning tasks. In code generation, PF-PPO achieves the state-of-the-art performance of 7-billion-parameter models on HumanEval (+7.9\%), MBPP (+0.7\%), and LeetCode Contest (+10.0\%) which is a more challenging benchmark created by us. In math reasoning, PF-PPO yields performance increase using different reward models and benchmarks (Ape210K and CMATH).
Code is available on \url{https://github.com/DtYXs/verl/tree/pf-ppo}.
\end{abstract}

\section{Introduction}
\label{sec:introduction}

Reinforcement Learning from Human Feedback (RLHF) is a key technique to align large language models (LLMs) with human values and preferences \citep{christiano2017deep,ziegler2019fine,ouyang2022training}.
RLHF has been proven to be an essential process for LLMs to produce more helpful, harmless, and honest responses \citep{bai2022training}.
Despite various non-RL algorithms such as DPO \citep{rafailov2024direct} are proposed,
state-of-the-art applications such as ChatGPT/GPT-4 \citep{openai2023gpt4}, Claude \citep{anthropic2023introducing}, and Gemini \citep{team2023gemini} adopt the RL algorithm (e.g., PPO) for policy optimization.
The key challenge of RLHF is the inaccuracy of the intermediate reward model.
While there are researchers investigate how to learn reliable reward models~\citep[see e.g.,][]{wang2024secrets}, we focus on how to learn better policy under the guidance of such inaccurate reward models.

\begin{figure}[t]
    \centering
    \includegraphics[width=\linewidth]{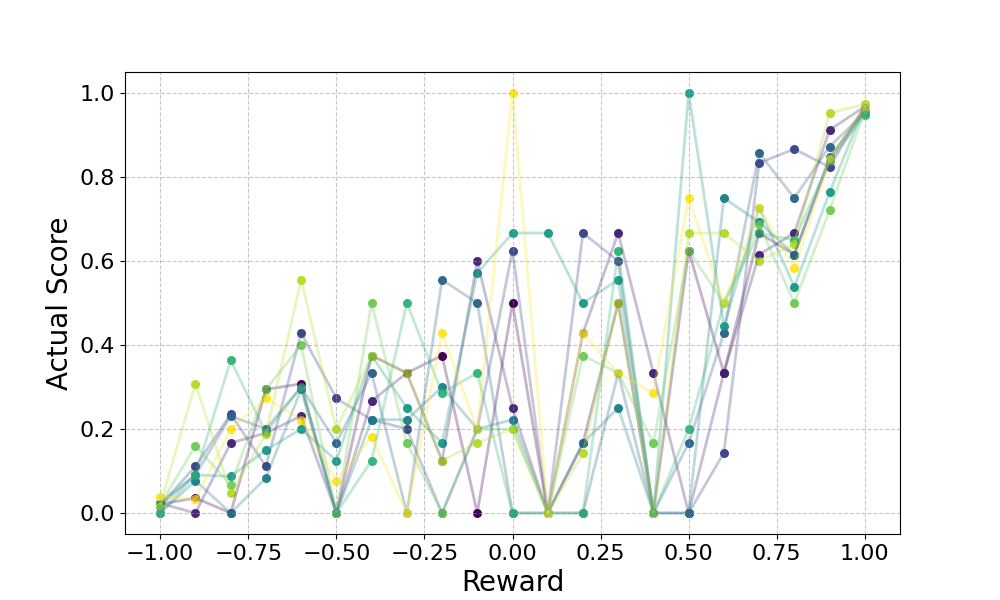}
    \caption{The reward model can be \emph{inaccurate}, i.e., the actual score of the response does not align well with the reward given by the reward model.
    However, the reward model in specific regions (e.g., when it gives rewards higher than 0.8) is more \emph{reliable}, i.e., the responses with similar rewards result in consistent performance.
    We use a fine-tuned policy to generate 10 responses for each of the 164 prompts in the HumanEval dataset and use a reward model trained with the common recipe to generate their rewards.
    We group the responses with similar rewards and calculate the average of their actual scores (i.e., the average correctness), indicating each group by one point.
    To evaluate the reliability of the reward model, we repeat the process ten times corresponding to the ten lines.
    }
    \label{fig:reward_model}
\end{figure}

We observe that, though the reward model gives inaccurate rewards in general, it can be more reliable in specific regions (e.g., when it gives high rewards) than the others.
The observation is based on the simple experiment:
We use a policy model fine-tuned for code generation to generate a set of responses for prompts in the HumanEval dataset.
Later, we score these responses using a reward model trained with the common recipe~\citep[see][and also Section 2]{ouyang2022training} and compare them with the actual scores. 
We find that, across different sets of samples, the reward model is more reliable when it gives high or low rewards than when it gives moderate rewards (cf. Figure~\ref{fig:reward_model}). 
This property also holds on other datasets and tasks and see Appendix~\ref{app:reward} for more experiment results and further discussion.
Considering that RLHF updates the policy solely based on the reward signal, this observation motivates us to filter out the samples with possibly unreliable rewards aiming to improve RLHF by increasing the signal-to-noise ratio on training samples.

Based on this motivation, we propose a simple modification to the standard PPO-based RLHF algorithm~\citep{ouyang2022training}, resulting in Policy Filtration for PPO (PF-PPO).
As in standard PPO, we generate $N$ samples for each prompt and score these samples using the reward model.
Then, we use a filtered subset of these samples in PF-PPO for subsequent policy training.
We design filtering strategies to improve the reliability of the reward model on the filtered samples by maximizing the coefficient of determination ($R^2$) between the rewards and actual scores on these filtered samples.
We show that the rewards of these filtered samples are more accurate, thus providing better training signal and improving the performance of the policy.
Our method is also connected with reject sampling that filters out responses with low rewards during inference to yield a better response.
Reject sampling is a simple but surprisingly strong inference-time strategy, whereas we adopt similar filtration in an RL algorithm.

Empirically, we show that PF-PPO can improve the performance of LLMs on the tasks where the complex logic makes the reward model inaccurate in general.
We conduct extensive ablation studies to validate the design of our algorithm.
In code generation, we illustrate the effectiveness of our algorithm by fine-tuning LLMs that achieves new sota on HumanEval and MBPP benchmarks across 7-billion-parameter LLMs.
We also create the LeetCode Contest benchmark that includes competition-level coding tasks for human experts and observe that PF-PPO results in even more significant improvement on this challenging benchmark. 
In math reasoning, we demonstrate that PF-PPO can improve the performance across different types of reward models.

\section{Related Work}
\label{sec:related_work}

\textbf{Limitation of reward model.}
The outcome of RLHF highly relies on the quality of the reward model.
Unfortunately, the reward model can hardly provide accurate scores due to 1) the mis-specified reward modeling to represent human preferences~\citep{lambert2023history,pitis2023failure}; 2) the presence of incorrect and ambiguous preferences in the dataset \citep{ouyang2022training,bai2022training}, and 3) the poor generalization ability of the reward model~\citep{mckinney2023fragility}.
The inaccuracy of reward model is attributed as one major cause of \emph{reward hacking} and \emph{hallucination} in LLMs \citep{kalai2024calibrated}.
While there are previous papers try to improve the accuracy of the reward model itself~\citep{wang2024secrets,coste2023reward,zhang2024improving},
the objective of our paper is to design a better RLHF algorithm in the face of inaccurate reward models. 
Moreover, \citet{bai2022training} also mentioned that using the output of the reward model directly in the RLHF process may not be a good choice. 
A possible solution is to penalize the outputs with low rewards more to improve the worst-case responses but they did not further implement this.

\textbf{Reject sampling.}
Reject sampling (or best-of-N sampling) is a popular and effective inference-time strategy to enhance the response of an LLM by generating $N$ responses and select the best one according to a reward model~\citep{nakano2021webgpt,cobbe2021training}.
This trick can yield good responses while keeping a tight KL constraint to the original policy.
Inspired by its effectiveness in inference, researchers also try to involve this trick in policy optimization.
For example, RAFT \citep{dong2023raft}, BOND \citep{sessa2024bond} and vBoN \citep{amini2024variational} learn a policy that distills the best-of-$N$ policy using supervised fine-tuning losses. 
In a boarder sense, the rank information of the $N$ samples can also be leveraged.
For example, RRHF \citep{yuan2023rrhf} and PRO \citep{song2024preference} train the policy using the combination of a ranking loss and a SFT loss (w.r.t. the best response) based on $N$ responses for each prompt.
However, these algorithms do not adopt an elaborate RL algorithm, while state-of-the-art language models adopts RL algorithms in alignment, benefiting from the generalization power of the reward model especially in reasoning tasks~\citep{ivison2024unpacking}.
Unlike these algorithms, we adopt the idea of reject sampling in the sampling phase of an RL algorithm instead of using supervised learning losses.



\textbf{RLHF algorithms in the face of inaccurate reward models.}
One key challenge in RLHF is the inaccuracy of reward model, which can lead to reward over-optimization~\citep{gao2023scaling,skalse2022defining,chaudhari2024rlhf}.
Optimization with a policy constraint (e.g., a KL divergence between the target policy and the reference policy) is a remedy frequently used in not only RL-based algorithms~\citep{ouyang2022training,wu2023pairwise,zhu2023fine} but also direct policy optimization algorithms~\citep{rafailov2024direct,zhao2023slic,liu2023statistical}.
Going beyond policy constraint, \citet{moskovitz2023confronting} only maximize rewards up to a threshold to avoid excessive deviation from a pre-trained policy.
In this paper, we not only rely on the policy constraint to optimize in the face of inaccurate rewards but also try to avoid using samples with unreliable rewards.


\section{Preliminary}
\label{sec:preliminary}

\textbf{Notations.}
We use $[a, b]$ to denote the set $\{a, a+1, \cdots, b\}$ and use $[b]$ as the shorthand for $[1, b]$.
We use $\oplus$ to denote the concatenation on tokens, and use $x_{a:b}$ as the shorthand for the concatenation $(x_a \oplus x_{a+1} \oplus \cdots \oplus x_b)$.
We use $c_i$ and $y_i$ to indicate the $i$-th token in the context $c$ (including task instruction, prompt, inputs, etc.) and the response $y$ respectively.

\textbf{MDP formulation.}
We adopt a Markov decision process (MDP) formulation for RLHF.
Specifically, language generation is formulated as an MDP $M=(\mathcal{S}, \mathcal{A}, P, R)$ with states $s\in\mathcal{S}$, actions $a\in\mathcal{A}$, transition probabilities $P\in\Delta(\mathcal{S})^{\mathcal{S}\times\mathcal{A}}$, and the next-state-based reward function $R: \mathcal{S} \to [0, 1]$. 
Given a context $c$ with $T_c$ tokens, on each step $t\in [T_c + 1, T]$\footnote{We fix the index of the terminal state to be the maximum length $T$. To adapt responses of different lengths, we left pad the context $c$.}, the language model $\pi_\theta(a_t|s_t)$ selects a token $a_t = y_{t-T_c}$ based on the state $s_t := (c_{1:T_c} \oplus y_{1:t-T_c-1})$.
Then, the language model enters the next state $s_{t+1} := (c_{1:T_c}\oplus y_{1:t-T_c})$ until the language model completes the response $y_{1:T-T_c}$.
For simplicity, we will also use contextual-bandit-style notations, e.g., we denote the language generation process as $y\sim\pi_\theta(\cdot|c)$.

\textbf{RLHF.}
Reinforcement learning with human feedback (RLHF) is an important process to address \emph{objective mismatch} between the next-token-prediction objective in pre-training and our expectation of LLMs to follow the instructions and assist humans to complete various tasks.
We briefly review the pipeline of RLHF.
\begin{itemize}[left=0pt]
\item \textbf{Supervised fine-tuning.}  
In the supervised fine-tuning (SFT) phase, a pre-trained LLM is fine-tuned with a high-quality supervised dataset collected for specific downstream tasks. Typically, the LLM is fine-tuned with a maximum likelihood loss, and we denote the output of this phase as $\pi^\text{SFT}$. While subsequent RLHF procedure is necessary for training high-quality LLMs, this phase alone can also yield an LLM that reasonably follows human instructions~\citep[see e.g.,][]{longpre2023flan}.
\item \textbf{Reward model learning.} 
In the reward model learning phase, we learn a reward model
$R_\phi(y | c) \in [-1, 1]$ parameterized by $\phi$ that scores the response $y$ to the context $c$ based on collected preference data $\mathcal{D}_\text{HF} := \{ (c, y^w, y^l) \}$ specifying that $y^w$ is a preferred response to $c$ than $y^l$.
The reward model is initialized by $\pi^\text{SFT}$ with an additional output layer.
A preference model links the reward model with the preference data, and Bradley-Terry model \citep{bradley1952rank} is a common choice:
\begin{equation}
    \mathbb{P}(y^w \succ y^l| c) = \sigma(R_\phi(y^w |c) - R_\phi(y^l|c)),
\end{equation}
where $\sigma$ is the sigmoid function.
The learning objective of reward model is to maximize the log-probability on preference data:
\begin{equation}
    \max_\phi \mathbb{E}_{(c, y_w, y_l) \sim \mathcal{D}_\text{HF}}
    \left[
    \log \mathbb{P}(y_w \succ y_l| c) 
    \right].
\end{equation}
\item \textbf{RL fine-tuning.}
In this stage, we fine-tune the language model $\pi_\theta$ to maximize the rewards given by the reward model with a policy constraint.
The optimization problem is formulated as
\begin{equation}
\label{eq:policy_optimization_objective}
\max_\theta \mathbb{E}_c \mathbb{E}_{y\sim \pi_\theta(\cdot|c)} 
\left[ 
r_\phi(y|c) - \beta D_\text{KL} (\pi_\theta(\cdot|c) || \pi^\text{SFT} (\cdot|c) )
\right].
\end{equation}
The second term prevents the learned policy deviating too much from the SFT model, and this is a popular technique to alleviate reward over-optimization~\citep{jaques2019way,stiennon2020learning}.
\end{itemize}

\textbf{PPO.}
Proximal policy optimization (PPO)~\citep{schulman2017proximal} is an RL algorithm that uses a clipped version of the policy gradient for more conservative and stable learning.
It becomes a standard algorithm for RL fine-tuning in RLHF that optimizes the modified (cumulative) reward 
\vspace{-10pt}
\begin{equation}
\label{eq:ppo_reward}
\begin{aligned}
    r_\phi(y|c) - \sum_{t=T_c+1}^T \beta \Big( & \log \pi_\theta(y_t|c\oplus y_{1:t-1}) \\
    & - \log \pi^\text{SFT}(y_t|c\oplus y_{1:t-1}) \Big)
\end{aligned}
\end{equation}
where the reward model gives sparse rewards and the policy constraint yields dense rewards.
PPO is an on-policy algorithm where the policy gradient is estimated based on the samples collected by the current policy $\pi_\theta$.

\begin{algorithm}
\caption{Proximal policy optimization (PPO)}
\label{alg:ppo}
\begin{algorithmic}
\FOR{iteration = $1, 2, \cdots$}
    \STATE Fill the buffer $\mathcal{B}$ with samples collected by the current language model $\pi_\theta$ 
    \STATE Update $\pi_\theta$ using PPO w.r.t. the cumulative reward defined in Equation~\eqref{eq:ppo_reward} based on $\mathcal{B}$
\ENDFOR
\end{algorithmic}
\end{algorithm}

\section{Methods}
\label{sec:methods}

Our method is motivated by the observation that the reward model is more reliable for the responses assigned with high/low rewards (cf. Figure~\ref{fig:reward_model}).
Consequently, we conjecture that, if we wrap the policy with proper filtration during policy optimization of RLHF, the reward model can avoid yielding unreliable rewards and thus give better signal to guide policy learning.

\textbf{Policy filtration.}
Given an unfiltered policy model $\pi_\theta(y| c)$ that generates responses $y$ to the context $c$, we denote the corresponding filtered policy as $\mu_\theta(y| c)$.
We consider a family of policy filtration, from which we can sample responses to the context $c$ as follows:
We first sample $N$ responses from $\pi_\theta(\cdot| c)$ and rank them by the reward model $R_\phi$, obtaining $y_1, \cdots, y_N$ with $R_\phi(y_1|c) \ge \cdots \ge R_\phi(y_N|c)$.
Then, given a weight vector $\mathbf{w}=(w_1, \cdots, w_N)$ satisfying $\sum_{i\in[N]} w_i = 1$, we sample a one-hot vector $\mathbf{z}=(z_1, \cdots, z_N)$ from the categorical distribution parameterized by $\mathbf{w}$ such that $\mathbb{P}[z_i=1] = w_i$.
At last, the filtered policy $\mu_\theta(\cdot|c)$ yields the response selected by $\mathbf{z}$ following $y=\sum_{i\in [N]} z_i y_i$.

We can define several filtered policies under this family.
Specifically, we obtain the best-of-$N$ (BoN), best-random (BR), and best-worst (BW) filtered policy by setting the weight vector as follows: 
\begin{equation}
\begin{aligned}
    &\mathbf{w}^\text{BoN} = (1, 0, \cdots, 0) \\
    &\mathbf{w}^\text{BR} =\left(\dfrac{1}{2}, \dfrac{1}{2(N-1)}, \cdots, \dfrac{1}{2(N-1)} \right) \\
    &\mathbf{w}^\text{BW} = \left( \dfrac{1}{2}, 0, \cdots, 0, \dfrac{1}{2} \right).
\end{aligned}
\end{equation}

\textbf{Training objective.}
Since our target is to learn a good filtered policy $\mu_\theta$, we consider the follow objective:
\begin{equation}
\label{eq:pf-ppo}
\max_\theta \mathbb{E}_c \mathbb{E}_{y\sim \textcolor{purple}{\mu_\theta(\cdot|c)}} 
\left[ 
r_\phi(y|c) - \beta D_\text{KL} (\textcolor{purple}{\mu_\theta(\cdot|x)} || \pi^\text{SFT} (\cdot|x) )
\right].
\end{equation}
In practice, use the samples collected by the unfiltered policy $\pi_\theta$ as if they were collected by $\textcolor{purple}{\mu_\theta}$ in the original PPO algorithm.
This leads to Policy Filtration Proximal Policy Optimization (PF-PPO) listed in Algorithm~\ref{alg:pf-ppo}, which is an algorithm that only modifies the sampling process of PPO.

\begin{algorithm}[t]
\caption{Policy Filtration Proximal policy Optimization (PF-PPO)}
\label{alg:pf-ppo}
\begin{algorithmic}
\FOR{iteration = $1, 2, \cdots$}
    \STATE Fill the buffer $\mathcal{B}$ with samples collected by the current language model $\mu_\theta$ 
    \STATE Update $\pi_\theta$ using PPO w.r.t. the cumulative reward defined in Equation~\eqref{eq:ppo_reward} based on $\mathcal{B}$
\ENDFOR
\end{algorithmic}
\end{algorithm}

\textbf{Weight choice.}
By defining different weight vectors $\mathbf{w}$, we can obtain different policy filtering strategies for PF-PPO.
Our objective is to choose a weight vector $\mathbf{w}$ such that the accuracy of the reward model on the responses generated by the filtered policies can be maximized.
To measure this accuracy, we choose a simple heuristic, the coefficient of determination (aka R-squared or $R^2$)~\citep{draper1998applied} between the rewards and the actual scores of the responses generated by the policy.
$R^2$ measures how well the actual scores can be predicted by the rewards with a linear model.
Specifically, given a set of responses $\{(c_i, y_i)\}$ sampled from the filtered policy $y_i \sim \mu_\theta(\cdot|c_i)$, we can collect the corresponding reward $R_i := R_\phi(y_i|c_i)$ and the actual score $s_i$.
Then, we fit a linear model $f$ to predict the actual score based on the reward and denote the predicted score as $\hat{s}_i=f(R_i)$.
The R-squared is calculated as $1-\frac{\sum_i(s_i - \hat{s}_i)^2}{\sum_i (s_i - \bar{s})^2}$ where $\bar{s}$ is the average of actual scores.
Since PF-PPO optimizes the policy based on the rewards on these responses, how well these rewards indicate the actual performance is closely related to the final performance of our algorithm.
We find $R^2$ well correlates with the final performance and can imply the level of reward over-optimization of the subsequent RLHF algorithm, therefore serving as a useful metrics to determine the weight vector used in PF-PPO.

To select a weight vector, we first checkpoint three policies $\pi_\theta$ collected from different stages of a standard RLHF process and collect responses using filtered policies $\mu_\theta$ in combination with different policy filtering strategies.
Then, we group the responses with similar rewards, record the average actual score and reward for each group, and calculate the $R^2$ by treating each group as a sample point.
We exam how different policy filtering strategies can improve the reliability of the rewards on the responses generated by the corresponding filtered policies.

We present the results in Table~\ref{tab:R2}.
We observe that best-random (BR) and best-worst (BW) can improve the reliability of the given reward model on sampled responses compared with unfiltered policy.
The BoN strategy does not improve the $R^2$, which indicates that learning a BoN filtered policy may not result in good performance in RL, although learning for a best-of-$N$ policy using supervised learning presents good performance~\citep{sessa2024bond}. 

\begin{table}[t]
    \centering
    \begin{tabular}{r|cccc}
        \hline
        Policy & No filter & BoN & BR & BW \\ 
        \hline 
        SFT & 0.886 & 0.454 & 0.922 & \textbf{0.952} \\
        Middle RLHF & 0.907 & 0.389 & 0.935 & \textbf{0.956} \\
        Final RLHF & 0.876& 0.431& 0.916& \textbf{0.946} \\
        \hline
    \end{tabular}
    \caption{The coefficient of determination ($R^2$) of the unfiltered policy $\pi_\theta$ and different filtered policies $\mu_\theta$ between the rewards given by the reward model and the actual scores. This metrics correlates well with the final performance (see Section~\ref{sec:experiments}) and helps us to determine the weight vector (or the policy filtering strategy) in our algorithm PF-PPO.
    We choose the SFT policy and the middle/final RLHF policy as the unfiltered policy $\pi_\theta$ respectively.
    }
    \label{tab:R2}
\end{table}

\section{Experiments}
\label{sec:experiments}

\subsection{Benchmarks}
We conduct experiments on two tasks where the quality of LLM responses can be precisely measured, code generation and math reasoning.
Specifically, we evaluate the algorithms using the following benchmarks.
 
\textbf{HumanEval benchmark and MBPP benchmark.}
HumanEval~\citep{chen2021evaluating} and MBPP~\citep{austin2021program} are two popular benchmarks for evaluating code LLMs. 
HumanEval includes 164 Python problems, each of which is associated with multiple test cases used to assess the correctness of generated code in a zero-shot setting. 
Similarly, MBPP includes 378 problems.

To train models for these two benchmarks, we select data from 75k Magicoder-OSS-instruct~\citep{wei2023magicoder} and 55k evol-codealpaca-v1~\citep{luo2023wizardcoder} to construct the SFT dataset, the reward model dataset, and the PPO query dataset.
For SFT, we use all the 130k training samples from Magicoder-OSS-instruct and evol-codealpaca-v1. 
For reward modeling, we curate 7k prompts from these 130k samples and generate five responses using the SFT model for each prompt. 
Following the methodology in \citet{pal2024smaug}, we select two responses with the maximum edit distance to create response pairs for each prompt. 
We use these 7k prompts with generated response pairs as the reward model dataset. 
For policy optimization, we curate 3k prompts from the 130k samples as the PPO query dataset.

\textbf{LeetCode contest benchmark.} 
To evaluate code LLMs on more challenging problems, we construct the LeetCode Contest benchmark. 
This benchmark includes competition-level problems designed for human, and therefore is more challenging since it requires human-level problem understanding and code generation skills. 
In this benchmark, we collect 160 problems from LeetCode weekly contests from July 2022 to January 2024. 
For each problem, we include 100 test cases to ensure the generated code is assessed thoroughly. 

To train models for this benchmark, we construct LeetCode training datasets comprising 1,000 problems collected from the LeetCode website.
For SFT, we use self-generated correct answers to create the SFT dataset following the methodology in \citet{setlur2024rl}. 
For reward modeling, we generate five responses using the SFT model for each of the 400 curated prompts and selected two responses for each prompt following the similar procedure as above. 
For policy optimization, we used the full 1,000 prompts as our PPO query dataset to train the code LLM.  

\textbf{Ape210K and CMATH benchmarks.}
Ape210K~\citep{zhao2020ape210k} and CMATH~\citep{wei2023cmath} are two popular Chinese benchmarks for elementary-school-level math reasoning tasks.
Ape210K contains 210k diverse math problems and we use a separate split of 5k problems for evaluation, following the practice in \citet{zhao2020ape210k}.
CMATH contains 1.7k math word problems sourcing from actual Chinese workbooks and exams.
We check the correctness of the answers using the automatic evaluation scripts provided in \citet{zhou2024leveraging}.
To train models for math reasoning, we use the training split of 200k problems from Ape210K.

\subsection{Implementation Details} 
We use deepseek-6.7B \citep{guo2024deepseek} and Qwen1.5-7B~\citep{qwen1.5} as our base model for code generation and math reasoning respectively. 
In the SFT phase, we train on the SFT dataset for 5 epochs with the learning rate $1\times 10^{-5}$, resulting in the SFT policy.
In the reward model training phase, we follow \citet{ouyang2022training} and train on our reward model dataset for 1 epoch with the learning rate  $1\times 10^{-5}$. 
In the PPO phase, we adopt the training tricks from the blog \citep{shen2024advanced}. 
Specifically, we adopt reward normalization and advantage normalization for stable training. 
In addition, we set the learning rate for the policy network as $5\times 10^{-7}$ and learning rate for the value network as $9\times 10^{-6}$.
In the PPO algorithm, we collect responses for the context in the PPO query dataset and iterate through this dataset for $5$ iterations (enough for convergence) and select the best checkpoints on evaluation set as the outcome policy.
For each collected context-response pair, we use it to accumulate loss and gradient for $3$ times on average.
We use full parameter fine-tuning in all the phases.
We provide the source code for all experiments in the supplementary.

\begin{table*}[t]
\centering
\begin{tabular}{llccc}
\toprule[1pt]
Family & Method & HumanEval & MBPP & LeetCode \\
\midrule
Supervised Fine-Tuning & SFT & 74.2 & 70.8 & 15.2  \\
& RAFT~\citep{dong2023raft} & 76.9 & 71.3 & 17.8 \\
& BOND~\citep{sessa2024bond} & 80.8 & 75.2 & 30.0 \\
\midrule
Direct Policy Optimization & DPO~\citep{rafailov2024direct} & 78.4 & 73.7 & 23.0 \\
& IPO~\citep{azar2024general} & 78.2 & 72.9 & 23.2 \\
& KTO~\citep{ethayarajh2024kto} & 77.9 & 72.5 & 22.4 \\
& Iterative-DPO~\citep{pang2024iterative} & 78.1 & 74.8 & 23.8 \\
\midrule
Reinforcement Learning & PPO-S~\citep{hu2024openrlhf} & 78.1 & 73.8 & 25.2 \\
& PPO-M~\citep[cf.][]{shao2024deepseekmath} & 80.2 & 75.0 & 29.8 \\
& PF-PPO (BoN) & 75.8 & 71.7 & 16.8 \\
& PF-PPO (BR) & \textbf{82.9} & \underline{75.9} & \textbf{33.0} \\
& PF-PPO (BW) & \underline{82.4} & \textbf{76.2} &  \underline{30.4} \\
\midrule
SOTA (7B models) & Magicoder~\citep{wei2023magicoder} & 76.8 & 75.7 & \\
\bottomrule[1pt]
\end{tabular}
\caption{\textbf{The performance of different algorithms on Code Generation.} 
We compare pass@1 of PF-PPO (our algorithm) against baseline methods. 
For each benchmark, we select the best score across 5 epochs for each method. 
The highest and the second highest scores on each benchmark are highlighted in \textbf{bold} and \underline{underline} respectively.
All experiments are based on the same code base for fair comparison, except for the scores reported by Magicoder which is the best 7B model so far.
}
\label{tab: performance comparison}
\end{table*}

\subsection{Experiment Results on Code Generation}

\textbf{Baselines.}
For code generation, we compare different variants of PF-PPO with reinforcement learning algorithms, supervised fine-tuning methods, and direct policy optimization methods.
We use greedy decoding during inference and pass@1~\citep{chen2021evaluating} as the performance metrics.
For fair comparison between different baselines, we re-implement these baselines with the same code base and the same datasets.
We also use the same reward model and the same SFT policy if applicable.

\emph{Supervised fine-tuning.}
Starting from deepseek-6.7B, we first fine-tune this policy on the SFT dataset. 
Other algorithms learn based on this SFT policy.
RAFT~\citep{dong2023raft} and BOND~\citep{sessa2024bond} train the policy to fit the best-of-$N$ (BoN) responses or the BoN policy via different supervised learning losses.
RAFT maximizes the log-probability of the BoN response, whereas
BOND minimizes a combination of the forward and backward KL divergence w.r.t. the BoN policy.
We set the coefficient to combine these two loss terms as $\beta_\text{BOND}=1.0$.
BOND is an iterative algorithm to fit the BoN policy based on the policy of the last iteration, and we train the policy for 4 iterations.

\emph{Direct policy optimization.} 
To implement direct policy optimization methods, we use our reward model dataset as the preference dataset required in these methods.
We implement DPO~\citep{rafailov2024direct}, IPO~\citep{azar2024general}, KTO~\citep{ethayarajh2024kto}, and iterative DPO~\citep{pang2024iterative}.
For iterative DPO, we train the DPO model for three iterations. 
For each iteration, we construct the preference dataset as follows:
The prompts are sampled from the reward model dataset and responses are generated by the trained DPO model from the previous iteration (if exists) or the previous SFT phase.  



\emph{Reinforcement Learning.} 
For standard RLHF, we use the implementation from OpenRLHF~\citep{hu2024openrlhf}, which incorporates several advanced PPO training techniques and has demonstrates strong performance on various benchmarks. 
We denote this baseline as PPO-S.
For our method PF-PPO, we implement three variants (BoN, BR, and BW) as introduced in the previous section.
Since PF-PPO collects multiple responses for a given prompt/context, we introduce another baseline called PPO-M (PPO with multiple responses) that uses all the $N$ responses for training without filtering.\footnote{
PPO-M can also be regarded as integrating GRPO~\citep{shao2024deepseekmath} into PPO, which has been adopted by Deepseek-V2~\citep{zhu2024deepseek} and Qwen2~\citep{yang2024qwen2}.}
The effective difference between PPO-S and PPO-M is that the buffer $\mathcal{B}$ in PPO-M contains more samples with the same context $c$ but with different responses $y$ which may provide detailed token-level instruction by comparing the responses corresponding to the same context.
Therefore, comparing with PPO-M can help us distinguish the effect of collecting multiple responses and that of filtering collected responses.
For fair comparison, we ensure that the computational costs of PF-PPO for each iteration is no larger than those of PPO-M and PPO-S, and we refer the readers to Appendix~\ref{app:computation_cost} for the detailed analysis on the computational efficiency of PPO-S, PPO-M, and PF-PPO.


\begin{figure*}[t!]
    \centering
    \begin{subfigure}[b]{0.47\linewidth}
        \centering
        \includegraphics[width=\textwidth]{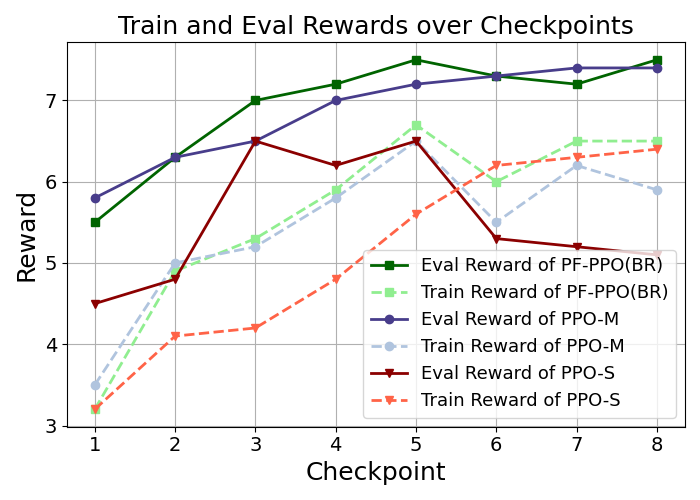}
    \end{subfigure}%
    \hfill
    \begin{subfigure}[b]{0.47\linewidth}
        \centering
        \includegraphics[width=\textwidth]{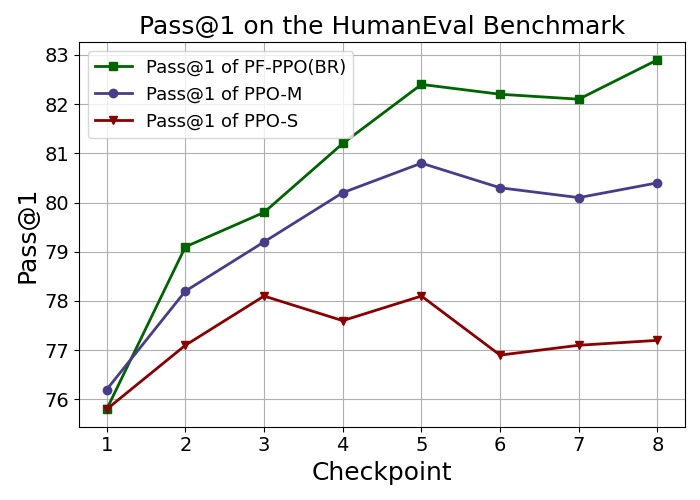}
    \end{subfigure}
    \caption{Left: The training and evaluation reward of PPO-S, PPO-M, and FP-PPO on HumanEval. 
    The training reward and the evaluation reward are evaluated on the samples generated by the filtered policy $\mu_\theta$ and the unfiltered policy $\pi_\theta$ respectively.
    Right: The pass@1 of PPO-S, PPO-M, and PF-PPO on the HumanEval benchmark.}
    \label{fig:details}
\end{figure*}

\textbf{Experiment results.}
We present the pass@1 results of different methods on the three benchmarks in Table~\ref{tab: performance comparison}. 
The experiment results show that PF-PPO (BR) and PF-PPO (BW) obtain the highest scores on these benchmarks, indicating the effectiveness of our method.
Furthermore, we have the following observations: 
\begin{itemize}[left=0pt]
    \item IPO and KTO (improved versions of DPO) do not outperform DPO when trained on properly selected datasets. This indicates that appropriate dataset construction can address the weaknesses of DPO found in previous papers, enabling DPO to achieve a performance comparable to its improved versions.
    \item PPO-based algorithms outperform SFT-based and DPO-based algorithms in general, demonstrating that PPO is superior to these algorithms on reasoning tasks. 
    We speculate that the good performance of PPO may stem from the generalization ability of the reward model and the value network used in PPO, which can be used to transform trajectory-level reward modeling to token-wise advantages and thus provides more fine-grained guidance. 
    Moreover, the gap between PPO-based algorithms and the others becomes larger on the more challenging LeetCode benchmark, which further highlights the advantage of RL on complex reasoning tasks
    \item BOND achieves the highest score among the baseline methods. It demonstrates that iterative best-of-$N$ (BoN) distillation is an effective alignment approach. We speculate that BOND also benefits from its ability to reduce learning on samples with unreliable rewards by selecting the best candidate from a set of $N$ samples.
    \item Motivated by the good performance of BOND, we implement PF-PPO (BoN) as a natural attempt to apply BoN to an RL-based algorithm.
    However, PF-PPO (BoN) results in poor performance.
    This indicates that compared with SFT methods that only need good samples, bad samples for the contrastive learning purposes are also important for RL-based methods.
    This explains the reason why PF-PPO (BR) and PF-PPO (BW) outperform PF-PPO (BoN).
    \item PF-PPO (BR) and PF-PPO (BW) outperform the others with a larger gap challenging LeetCode tasks. 
    We find that the accuracy of the reward model decreases on this benchmark since it is more difficult for the reward model to distinguish whether one response is better than another, especially when both responses contain errors. 
    This decreases the reliability of the reward model in the moderate reward region (cf. Figure~\ref{fig:reward_model}). 
    Consequently, PF-PPO (BR) and PF-PPO (BW) can improve the performance in these complex reasoning tasks by avoiding learning on unreliable rewards.
\end{itemize}  

\textbf{Training curves.}
To provide a comprehensive view of the three algorithms, we show the details of the training process.

We first present the training curves of PPO-S, PPO-M, and PF-PPO in Figure \ref{fig:details} (left). 
The training rewards are evaluated on the samples collected by the filtered policy $\mu_\theta$ and the evaluation rewards are calculated on the unfiltered policy $\pi_\theta$.
We observe that both the training reward and evaluation reward of PPO-M and PF-PPO surpass those of PPO-S. 
This indicates that sampling multiple responses from a context enhances the performance of the RLHF method, consistent with the findings in \citet{shao2024deepseekmath}. 
Moreover, in terms of optimizing reward for the same given reward model, FP-PPO achieves a higher or equal reward compared with PPO-S and PPO-M, which indicates that the approximation made in the FP-PPO (i.e., optimizing the unfiltered policy $\pi_\theta$ as if it were the filtered policy $\mu_\theta$) does not induce negative effect on its capability of optimizing the reward.


We also show the pass@1 results of different algorithms in Figure~\ref{fig:details} (right).
We observe that, while PF-PPO achieves a similar reward to that of PPO-M, the pass@1 result of PF-PPO exceeds that of PPO-M significantly.
This results from the fact that PF-PPO optimizes on the reliable region of the reward model and thus alleviate the reward over-optimization issue.


\subsection{Alternative Policy Filtering Strategies}

PF-PPO modifies the sampling procedure of standard PPO by sampling $N$ responses and randomly filtering responses based on their ranks.
In this part, we consider other alternatives to filter by threshold or down-weight the responses with unreliable rewards in the sampling procedure.
\begin{itemize}[left=0pt]
    \item \textbf{Filtering based on reward thresholds.} 
    Given a reward model, we can filter the responses based on their rewards using specified threshold. 
    This results in three strategies, 
    \emph{PPO-top} that only keeps the top samples whose rewards exceeding a certain threshold,
    \emph{PPO-top-random} that keeps also keeps random samples with 50\% probability,
    and \emph{PPO-top-bottom} that keeps top samples and bottom samples whose rewards are below another specified threshold. 
    These strategies can be regarded as the threshold version of PF-PPO (BoN), PF-PPO (BR) and PF-PPO (BW) respectively.
    The thresholds are tuned coarsely to achieve good results on a separate validation set.
    \item \textbf{Filtering based on reward reweighting.}
    Compared with the above strategies that use thresholds, we consider a softer version that adjusts the sample weights based on their rewards, aiming at down-weight the samples with moderate and possibly unreliable rewards.
    Specifically, we increase the sample weight of the responses with rewards in the reliable region and decrease the sample weight otherwise. 
    To achieve this goal, given a reward model $R_\phi$ that returns rewards in the range $[-1, 1]$, we assign the weight for the sample $(c, y)$ proportional to $|R_\phi(y|c)|^k$ and collect samples with these weights from the buffer $\mathcal{B}$ to train the policy network and the value network. 
    We denote these strategies as \emph{PPO-pow-$k$}.
\end{itemize}

\begin{table}
\small
\centering
\begin{tabular}{L{2cm}C{1.5cm}C{1.5cm}C{1.5cm}}
\toprule[1pt]
Policy filtering strategies & pass@1 on Human Eval & pass@1 on MBPP & $R^2$ based on SFT policy \\
\midrule
PPO             &  \gradientcell{78.1}{75.8}{82.9}{cyan}{purple}{30}    & \gradientcell{73.8}{71.2}{76.5}{cyan}{purple}{30} & \gradientcell{0.782}{0.454}{0.952}{cyan}{purple}{30} \\
PPO-M           &  \gradientcell{80.8}{75.8}{82.9}{cyan}{purple}{30}    & \gradientcell{75.0}{71.2}{76.5}{cyan}{purple}{30} & \gradientcell{0.886}{0.454}{0.952}{cyan}{purple}{30} \\
PF-PPO (BoN)    & \gradientcell{75.8}{75.8}{82.9}{cyan}{purple}{30}     & \gradientcell{71.7}{71.2}{76.5}{cyan}{purple}{30} & \gradientcell{0.454}{0.454}{0.952}{cyan}{purple}{30} \\
PF-PPO (BR)     & \gradientcell{82.9}{75.8}{82.9}{cyan}{purple}{30}     & \gradientcell{75.9}{71.2}{76.5}{cyan}{purple}{30} & \gradientcell{0.841}{0.454}{0.952}{cyan}{purple}{30} \\
PF-PPO (BW)     & \gradientcell{82.4}{75.8}{82.9}{cyan}{purple}{30}     & \gradientcell{76.2}{71.2}{76.5}{cyan}{purple}{30} & \gradientcell{0.952}{0.454}{0.952}{cyan}{purple}{30} \\
PPO-top         & \gradientcell{80.5}{75.8}{82.9}{cyan}{purple}{30}     & \gradientcell{71.2}{71.2}{76.5}{cyan}{purple}{30} & \gradientcell{0.621}{0.454}{0.952}{cyan}{purple}{30} \\
PPO-top-rand  & \gradientcell{81.9}{75.8}{82.9}{cyan}{purple}{30}     & \gradientcell{75.3}{71.2}{76.5}{cyan}{purple}{30} & \gradientcell{0.889}{0.454}{0.952}{cyan}{purple}{30} \\
PPO-top-bott  & \gradientcell{81.7}{75.8}{82.9}{cyan}{purple}{30}     & \gradientcell{75.4}{71.2}{76.5}{cyan}{purple}{30} & \gradientcell{0.927}{0.454}{0.952}{cyan}{purple}{30} \\
PPO-pow-1       & \gradientcell{81.0}{75.8}{82.9}{cyan}{purple}{30}     & \gradientcell{74.2}{71.2}{76.5}{cyan}{purple}{30} & \gradientcell{0.926}{0.454}{0.952}{cyan}{purple}{30} \\
PPO-pow-2       & \gradientcell{81.3}{75.8}{82.9}{cyan}{purple}{30}     & \gradientcell{75.4}{71.2}{76.5}{cyan}{purple}{30} & \gradientcell{0.939}{0.454}{0.952}{cyan}{purple}{30} \\
PPO-pow-3       & \gradientcell{81.9}{75.8}{82.9}{cyan}{purple}{30}     & \gradientcell{76.5}{71.2}{76.5}{cyan}{purple}{30} & \gradientcell{0.946}{0.454}{0.952}{cyan}{purple}{30} \\
\bottomrule[1pt]
\end{tabular}
\caption{
The comparison on the pass@1 results of different policy filtering strategies on HumanEval and their corresponding $R^2$ based on the SFT policy.
The background are colored based on their values with blue and red indicating the minimum and the maximum respectively.
}
\label{tab: ablation study}
\vspace{-5pt}
\end{table}

A question then arises: how to choose a policy filtering strategy from these strategies?
To answer this question, we propose to calculate the $R^2$ between the rewards and the actual scores on the samples collected by different strategies, and then choose a strategy with good results on this metrics.
We can use the SFT policy as the unfiltered policy and calculate $R^2$ as described in Section~\ref{sec:methods}.
Since the SFT policy is obtained prior to the PPO training phase, this metric can be used to predict the results of different filtering strategies before actually conduct costly PPO training.



We compare theses strategies on HumanEval and present the performance of different policy filtering strategies and their corresponding $R^2$ in Table~\ref{tab: ablation study}.
We make the following observations:
First, the $R^2$ of different strategies positively correlate with their performance in general, indicating $R^2$ can serve as a tool to predict the performance of different policy filtering strategies.
Second, different policy filtering strategies (except for BoN versions) improve the performance of the base PPO algorithms. 
This indicates that filtering samples with unreliable rewards can increase the signal-to-noise ratio of the reward model feedback and thus improve the performance. 
Third, PF-PPO strategies (which are rank-based) outperforms other strategies (which are threshold-based or reweighting-based).
This may due to the fact that rank-based strategies are more robust to the reward distribution of the given reward model.

\textbf{Discussion.}
The performance of different policy filtering strategies may vary across different tasks, different reward models, and different base models.
Therefore, although we find that PF-PPO (BR) and PF-PPO (BW) are the best strategies in our setting, other policy filtering strategies may be a better choice in other settings. 
Therefore, a more practical procedure should be first calculate the $R^2$ using the given reward model and the corresponding SFT policy on the specific task and select candidate policy filtering strategies. 
Note that $R^2$ is not a perfect tool to select policy filtering strategies and we leave seeking for better predictive metrics as a future research direction.


\subsection{Experiment Results on Math Reasoning}
\label{sec:exp_math}

To evaluate the effectiveness of PF-PPO in other domains and different types of reward models, we applied PF-PPO to solve math problems.
We consider three types of reward models: the original reward model (ORM) that is trained on preference datasets using a Bradley–Terry model~\citep{bradley1952rank}, an oracle model (Oracle) that extracts the final answer from the response and compares it with the ground truth, and a combined reward model (CRM) that integrates the above two models, similar to the approach used in Qwen-Math~\citep{yang2024qwen2}. 
We compare PF-PPO (BR) to PPO-S using these reward models.

\begin{table}[t]
    \centering
    \begin{tabular}{lrr}
    \toprule
                    & Ape210K & CMATH   \\ \midrule
    PPO-S + ORM     & 84.1    & 92.3    \\
    PF-PPO + ORM    & \textbf{86.2} & \textbf{95.1}    \\
    PPO-S + Oracle      & 82.1    & 90.8 \\
    PF-PPO + Oracle     & \textbf{83.8} & \textbf{91.2} \\
    PPO-S + CRM     & 83.9  & 93.1  \\
    PF-PPO + CRM    & \textbf{84.3} & \textbf{94.2} \\ \bottomrule
    \end{tabular}
    \caption{Comparison between PF-PPO and PPO-S on two math benchmarks (Ape210K and CMATH) using three different reward functions (the original reward model, the oracle model, and the combined reward model). Better results for each reward model is highlighted in \textbf{bold}.
    }
    \label{tab:math}
\end{table}

\begin{table*}[t]
    \centering
    \begin{tabular}{lrrr}
    \toprule
    Task & Evaluation Set Size & BO1 Accuracy (\%)	& BO5 Accuracy (\%)  \\ 
    \midrule
Logic Reasoning     & 1,203      & \textbf{48.9 (+2.3)}   & \textbf{63.8 (+2.8)} \\ 
Math                & 1,759      & \textbf{69.7 (+1.1)}	& \textbf{79.9 (+2.3)} \\
Code                & 3,933      & 55.8 (-0.2)	& 67.4 (+0.1) \\
STEM                & 4,466      & 54.7 (-0.1)   & 63.1 (+0.1) \\
Complex Tasks       & 2,990      & \textbf{9.5 (+1.0)}	& \textbf{14.9 (+0.6)} \\
Instruction Following & 1,525    & \textbf{49.6 (+1.7)}	& \textbf{59.8 (+1.8)} \\
Knowledge           & 775       & \textbf{47.3 (+1.9)}	& \textbf{58.3 (+1.8)} \\
Language Understanding & 680    & \textbf{63.8 (+1.6)}	& \textbf{68.4 (+3.8)} \\
    \bottomrule
    \end{tabular}
    \caption{
    Improvement of PF-PPO compared with PPO-S on a wide range of tasks. We present the best-of-1 (BO1) and best-of-5 (BO5) accuracies for PF-PPO and the accuracy improvement of PF-PPO compared to PPO-S. The results with significant improvement are highlighted in \textbf{bold}.
    }
    \label{tab:doubao}
\end{table*}

We can observe that PF-PPO consistently outperforms the PPO algorithm on these two benchmarks across different reward models. 
In addition, the experiment results indicate that even if we can have access to the ground truth, using the oracle as the reward function does not perform as well as using a reward model (either the original reward model or the combined model).
This finding is consistent with experiment results in Qwen-Math~\citep{yang2024qwen2} and Deepseek-Math~\citep{shao2024deepseekmath}.





\subsection{Experiment Results on Wider Range of Tasks}
\label{sec:other_tasks}

To further validate the broader effectiveness of our method, we conducted experiments across diverse domains using Doubao-25k (policy and reward model backbone). 
Tasks included logic reasoning, math, code generation, STEM problems, complex tasks, instruction following, knowledge QA, and language understanding. 
Each task has distinct evaluation sets and verifiers to assess response correctness. 
We consider the multi-task scenario where one model is trained to complete various tasks.
We present the results (accuracy improvement over vanilla PPO) in Table~\ref{tab:doubao}.
We highlight statistically significant changes (exceeding $\pm 0.5 \%$, based on test case counts) in bold. 
These results demonstrate PF-PPO’s consistent effectiveness across tasks.

\section{Conclusion}
\label{sec:conclusion}
In this paper, we propose a new reinforcement learning with human feedback (RLHF) method, \textbf{Policy Filtration for Proximal Policy Optimization (PF-PPO)}, aimed at mitigating the adverse effects of reward noise. 
When training the reward model using the Bradley-Terry approach, the reward signal is generally more reliable in the high or low reward regions but less reliable in the moderate reward regions. 
Motivated by this observation, we adopt a rank-based method to selectively use sample from these reliable regions more in PPO to improve the quality of the signal provided by the reward model.
We conduct comprehensive experiments to demonstrate that PF-PPO consistently outperforms existing baselines.
Additionally, we analyze PF-PPO, standard PPO, and PPO with multiple responses in details and show that filtering samples with unreliable rewards can improve the performance of the outcome policy.



\section*{Acknowledgment}
The work was partially supported by The State Key Laboratory of Novel Software Technology (KFKT2024A03).

\section*{Impact Statement}

This paper presents work whose goal is to advance the field of 
Machine Learning. There are many potential societal consequences 
of our work, none which we feel must be specifically highlighted here.

\bibliography{main}

\begin{thebibliography}{56}
\providecommand{\natexlab}[1]{#1}
\providecommand{\url}[1]{\texttt{#1}}
\expandafter\ifx\csname urlstyle\endcsname\relax
  \providecommand{\doi}[1]{doi: #1}\else
  \providecommand{\doi}{doi: \begingroup \urlstyle{rm}\Url}\fi

\bibitem[Amini et~al.(2024)Amini, Vieira, and Cotterell]{amini2024variational}
Amini, A., Vieira, T., and Cotterell, R.
\newblock Variational best-of-n alignment.
\newblock \emph{arXiv preprint arXiv:2407.06057}, 2024.

\bibitem[Anthropic(2023)]{anthropic2023introducing}
Anthropic, A.
\newblock Introducing claude, 2023.
\newblock URL \url{https://www.anthropic.com/news/introducing-claude}.

\bibitem[Austin et~al.(2021)Austin, Odena, Nye, Bosma, Michalewski, Dohan, Jiang, Cai, Terry, Le, et~al.]{austin2021program}
Austin, J., Odena, A., Nye, M., Bosma, M., Michalewski, H., Dohan, D., Jiang, E., Cai, C., Terry, M., Le, Q., et~al.
\newblock Program synthesis with large language models.
\newblock \emph{arXiv preprint arXiv:2108.07732}, 2021.

\bibitem[Azar et~al.(2024)Azar, Guo, Piot, Munos, Rowland, Valko, and Calandriello]{azar2024general}
Azar, M.~G., Guo, Z.~D., Piot, B., Munos, R., Rowland, M., Valko, M., and Calandriello, D.
\newblock A general theoretical paradigm to understand learning from human preferences.
\newblock In \emph{International Conference on Artificial Intelligence and Statistics}, pp.\  4447--4455. PMLR, 2024.

\bibitem[Bai et~al.(2022)Bai, Jones, Ndousse, Askell, Chen, DasSarma, Drain, Fort, Ganguli, Henighan, et~al.]{bai2022training}
Bai, Y., Jones, A., Ndousse, K., Askell, A., Chen, A., DasSarma, N., Drain, D., Fort, S., Ganguli, D., Henighan, T., et~al.
\newblock Training a helpful and harmless assistant with reinforcement learning from human feedback.
\newblock \emph{arXiv preprint arXiv:2204.05862}, 2022.

\bibitem[Bradley \& Terry(1952)Bradley and Terry]{bradley1952rank}
Bradley, R.~A. and Terry, M.~E.
\newblock Rank analysis of incomplete block designs: I. the method of paired comparisons.
\newblock \emph{Biometrika}, 39\penalty0 (3/4):\penalty0 324--345, 1952.

\bibitem[Chaudhari et~al.(2024)Chaudhari, Aggarwal, Murahari, Rajpurohit, Kalyan, Narasimhan, Deshpande, and da~Silva]{chaudhari2024rlhf}
Chaudhari, S., Aggarwal, P., Murahari, V., Rajpurohit, T., Kalyan, A., Narasimhan, K., Deshpande, A., and da~Silva, B.~C.
\newblock Rlhf deciphered: A critical analysis of reinforcement learning from human feedback for llms.
\newblock \emph{arXiv preprint arXiv:2404.08555}, 2024.

\bibitem[Chen et~al.(2021)Chen, Tworek, Jun, Yuan, Pinto, Kaplan, Edwards, Burda, Joseph, Brockman, et~al.]{chen2021evaluating}
Chen, M., Tworek, J., Jun, H., Yuan, Q., Pinto, H. P. d.~O., Kaplan, J., Edwards, H., Burda, Y., Joseph, N., Brockman, G., et~al.
\newblock Evaluating large language models trained on code.
\newblock \emph{arXiv preprint arXiv:2107.03374}, 2021.

\bibitem[Christiano et~al.(2017)Christiano, Leike, Brown, Martic, Legg, and Amodei]{christiano2017deep}
Christiano, P.~F., Leike, J., Brown, T., Martic, M., Legg, S., and Amodei, D.
\newblock Deep reinforcement learning from human preferences.
\newblock \emph{Advances in neural information processing systems}, 30, 2017.

\bibitem[Cobbe et~al.(2021)Cobbe, Kosaraju, Bavarian, Chen, Jun, Kaiser, Plappert, Tworek, Hilton, Nakano, et~al.]{cobbe2021training}
Cobbe, K., Kosaraju, V., Bavarian, M., Chen, M., Jun, H., Kaiser, L., Plappert, M., Tworek, J., Hilton, J., Nakano, R., et~al.
\newblock Training verifiers to solve math word problems.
\newblock \emph{arXiv preprint arXiv:2110.14168}, 2021.

\bibitem[Coste et~al.(2023)Coste, Anwar, Kirk, and Krueger]{coste2023reward}
Coste, T., Anwar, U., Kirk, R., and Krueger, D.
\newblock Reward model ensembles help mitigate overoptimization.
\newblock \emph{arXiv preprint arXiv:2310.02743}, 2023.

\bibitem[Dong et~al.(2023)Dong, Xiong, Goyal, Zhang, Chow, Pan, Diao, Zhang, Shum, and Zhang]{dong2023raft}
Dong, H., Xiong, W., Goyal, D., Zhang, Y., Chow, W., Pan, R., Diao, S., Zhang, J., Shum, K., and Zhang, T.
\newblock Raft: Reward ranked finetuning for generative foundation model alignment.
\newblock \emph{arXiv preprint arXiv:2304.06767}, 2023.

\bibitem[Draper(1998)]{draper1998applied}
Draper, N.
\newblock \emph{Applied regression analysis}.
\newblock McGraw-Hill. Inc, 1998.

\bibitem[Ethayarajh et~al.(2024)Ethayarajh, Xu, Muennighoff, Jurafsky, and Kiela]{ethayarajh2024kto}
Ethayarajh, K., Xu, W., Muennighoff, N., Jurafsky, D., and Kiela, D.
\newblock Kto: Model alignment as prospect theoretic optimization.
\newblock \emph{arXiv preprint arXiv:2402.01306}, 2024.

\bibitem[Gao et~al.(2023)Gao, Schulman, and Hilton]{gao2023scaling}
Gao, L., Schulman, J., and Hilton, J.
\newblock Scaling laws for reward model overoptimization.
\newblock In \emph{International Conference on Machine Learning}, pp.\  10835--10866. PMLR, 2023.

\bibitem[Guo et~al.(2024)Guo, Zhu, Yang, Xie, Dong, Zhang, Chen, Bi, Wu, Li, et~al.]{guo2024deepseek}
Guo, D., Zhu, Q., Yang, D., Xie, Z., Dong, K., Zhang, W., Chen, G., Bi, X., Wu, Y., Li, Y., et~al.
\newblock Deepseek-coder: When the large language model meets programming--the rise of code intelligence.
\newblock \emph{arXiv preprint arXiv:2401.14196}, 2024.

\bibitem[Hu et~al.(2024)Hu, Wu, Wang, Zhang, Cao, et~al.]{hu2024openrlhf}
Hu, J., Wu, X., Wang, W., Zhang, D., Cao, Y., et~al.
\newblock Openrlhf: An easy-to-use, scalable and high-performance rlhf framework.
\newblock \emph{arXiv preprint arXiv:2405.11143}, 2024.

\bibitem[Ivison et~al.(2024)Ivison, Wang, Liu, Wu, Pyatkin, Lambert, Smith, Choi, and Hajishirzi]{ivison2024unpacking}
Ivison, H., Wang, Y., Liu, J., Wu, Z., Pyatkin, V., Lambert, N., Smith, N.~A., Choi, Y., and Hajishirzi, H.
\newblock Unpacking dpo and ppo: Disentangling best practices for learning from preference feedback.
\newblock \emph{arXiv preprint arXiv:2406.09279}, 2024.

\bibitem[Jaques et~al.(2019)Jaques, Ghandeharioun, Shen, Ferguson, Lapedriza, Jones, Gu, and Picard]{jaques2019way}
Jaques, N., Ghandeharioun, A., Shen, J.~H., Ferguson, C., Lapedriza, A., Jones, N., Gu, S., and Picard, R.
\newblock Way off-policy batch deep reinforcement learning of implicit human preferences in dialog.
\newblock \emph{arXiv preprint arXiv:1907.00456}, 2019.

\bibitem[Kalai \& Vempala(2024)Kalai and Vempala]{kalai2024calibrated}
Kalai, A.~T. and Vempala, S.~S.
\newblock Calibrated language models must hallucinate.
\newblock In \emph{Proceedings of the 56th Annual ACM Symposium on Theory of Computing}, pp.\  160--171, 2024.

\bibitem[Lambert et~al.(2023)Lambert, Krendl~Gilbert, and Zick]{lambert2023history}
Lambert, N., Krendl~Gilbert, T., and Zick, T.
\newblock The history and risks of reinforcement learning and human feedback.
\newblock \emph{arXiv e-prints}, pp.\  arXiv--2310, 2023.

\bibitem[Liu et~al.(2023)Liu, Zhao, Joshi, Khalman, Saleh, Liu, and Liu]{liu2023statistical}
Liu, T., Zhao, Y., Joshi, R., Khalman, M., Saleh, M., Liu, P.~J., and Liu, J.
\newblock Statistical rejection sampling improves preference optimization.
\newblock \emph{arXiv preprint arXiv:2309.06657}, 2023.

\bibitem[Longpre et~al.(2023)Longpre, Hou, Vu, Webson, Chung, Tay, Zhou, Le, Zoph, Wei, et~al.]{longpre2023flan}
Longpre, S., Hou, L., Vu, T., Webson, A., Chung, H.~W., Tay, Y., Zhou, D., Le, Q.~V., Zoph, B., Wei, J., et~al.
\newblock The flan collection: Designing data and methods for effective instruction tuning.
\newblock In \emph{International Conference on Machine Learning}, pp.\  22631--22648. PMLR, 2023.

\bibitem[Luo et~al.(2023)Luo, Xu, Zhao, Sun, Geng, Hu, Tao, Ma, Lin, and Jiang]{luo2023wizardcoder}
Luo, Z., Xu, C., Zhao, P., Sun, Q., Geng, X., Hu, W., Tao, C., Ma, J., Lin, Q., and Jiang, D.
\newblock Wizardcoder: Empowering code large language models with evol-instruct.
\newblock \emph{arXiv preprint arXiv:2306.08568}, 2023.

\bibitem[McKinney et~al.(2023)McKinney, Duan, Krueger, and Gleave]{mckinney2023fragility}
McKinney, L., Duan, Y., Krueger, D., and Gleave, A.
\newblock On the fragility of learned reward functions.
\newblock \emph{arXiv preprint arXiv:2301.03652}, 2023.

\bibitem[Moskovitz et~al.(2023)Moskovitz, Singh, Strouse, Sandholm, Salakhutdinov, Dragan, and McAleer]{moskovitz2023confronting}
Moskovitz, T., Singh, A.~K., Strouse, D., Sandholm, T., Salakhutdinov, R., Dragan, A.~D., and McAleer, S.
\newblock Confronting reward model overoptimization with constrained rlhf.
\newblock \emph{arXiv preprint arXiv:2310.04373}, 2023.

\bibitem[Nakano et~al.(2021)Nakano, Hilton, Balaji, Wu, Ouyang, Kim, Hesse, Jain, Kosaraju, Saunders, et~al.]{nakano2021webgpt}
Nakano, R., Hilton, J., Balaji, S., Wu, J., Ouyang, L., Kim, C., Hesse, C., Jain, S., Kosaraju, V., Saunders, W., et~al.
\newblock Webgpt: Browser-assisted question-answering with human feedback.
\newblock \emph{arXiv preprint arXiv:2112.09332}, 2021.

\bibitem[OpenAI(2023)]{openai2023gpt4}
OpenAI.
\newblock Gpt-4 technical report.
\newblock \emph{arXiv preprint arXiv:2303.08774}, 2023.

\bibitem[Ouyang et~al.(2022)Ouyang, Wu, Jiang, Almeida, Wainwright, Mishkin, Zhang, Agarwal, Slama, Ray, et~al.]{ouyang2022training}
Ouyang, L., Wu, J., Jiang, X., Almeida, D., Wainwright, C., Mishkin, P., Zhang, C., Agarwal, S., Slama, K., Ray, A., et~al.
\newblock Training language models to follow instructions with human feedback.
\newblock \emph{Advances in Neural Information Processing Systems}, 35:\penalty0 27730--27744, 2022.

\bibitem[Pal et~al.(2024)Pal, Karkhanis, Dooley, Roberts, Naidu, and White]{pal2024smaug}
Pal, A., Karkhanis, D., Dooley, S., Roberts, M., Naidu, S., and White, C.
\newblock Smaug: Fixing failure modes of preference optimisation with dpo-positive.
\newblock \emph{arXiv preprint arXiv:2402.13228}, 2024.

\bibitem[Pang et~al.(2024)Pang, Yuan, Cho, He, Sukhbaatar, and Weston]{pang2024iterative}
Pang, R.~Y., Yuan, W., Cho, K., He, H., Sukhbaatar, S., and Weston, J.
\newblock Iterative reasoning preference optimization.
\newblock \emph{arXiv preprint arXiv:2404.19733}, 2024.

\bibitem[Pitis(2023)]{pitis2023failure}
Pitis, S.
\newblock Failure modes of learning reward models for llms and other sequence models.
\newblock In \emph{ICML 2023 Workshop The Many Facets of Preference-Based Learning}, 2023.

\bibitem[Rafailov et~al.(2024)Rafailov, Sharma, Mitchell, Manning, Ermon, and Finn]{rafailov2024direct}
Rafailov, R., Sharma, A., Mitchell, E., Manning, C.~D., Ermon, S., and Finn, C.
\newblock Direct preference optimization: Your language model is secretly a reward model.
\newblock \emph{Advances in Neural Information Processing Systems}, 36, 2024.

\bibitem[Schulman et~al.(2017)Schulman, Wolski, Dhariwal, Radford, and Klimov]{schulman2017proximal}
Schulman, J., Wolski, F., Dhariwal, P., Radford, A., and Klimov, O.
\newblock Proximal policy optimization algorithms.
\newblock \emph{arXiv preprint arXiv:1707.06347}, 2017.

\bibitem[Sessa et~al.(2024)Sessa, Dadashi, Hussenot, Ferret, Vieillard, Ram{\'e}, Shariari, Perrin, Friesen, Cideron, et~al.]{sessa2024bond}
Sessa, P.~G., Dadashi, R., Hussenot, L., Ferret, J., Vieillard, N., Ram{\'e}, A., Shariari, B., Perrin, S., Friesen, A., Cideron, G., et~al.
\newblock Bond: Aligning llms with best-of-n distillation.
\newblock \emph{arXiv preprint arXiv:2407.14622}, 2024.

\bibitem[Setlur et~al.(2024)Setlur, Garg, Geng, Garg, Smith, and Kumar]{setlur2024rl}
Setlur, A., Garg, S., Geng, X., Garg, N., Smith, V., and Kumar, A.
\newblock Rl on incorrect synthetic data scales the efficiency of llm math reasoning by eight-fold.
\newblock \emph{arXiv preprint arXiv:2406.14532}, 2024.

\bibitem[Shao et~al.(2024)Shao, Wang, Zhu, Xu, Song, Zhang, Li, Wu, and Guo]{shao2024deepseekmath}
Shao, Z., Wang, P., Zhu, Q., Xu, R., Song, J., Zhang, M., Li, Y., Wu, Y., and Guo, D.
\newblock Deepseekmath: Pushing the limits of mathematical reasoning in open language models.
\newblock \emph{arXiv preprint arXiv:2402.03300}, 2024.

\bibitem[Shen et~al.(2024)Shen, Hu, Zhao, He, and Chen]{shen2024advanced}
Shen, W., Hu, J., Zhao, P., He, X., and Chen, L.
\newblock Advanced tricks for training large language models with proximal policy optimization.
\newblock \url{https://difficult-link-dd7.notion.site/eb7b2d1891f44b3a84e7396d19d39e6f}, 2024.
\newblock Notion Blog.

\bibitem[Skalse et~al.(2022)Skalse, Howe, Krasheninnikov, and Krueger]{skalse2022defining}
Skalse, J., Howe, N., Krasheninnikov, D., and Krueger, D.
\newblock Defining and characterizing reward gaming.
\newblock \emph{Advances in Neural Information Processing Systems}, 35:\penalty0 9460--9471, 2022.

\bibitem[Song et~al.(2024)Song, Yu, Li, Yu, Huang, Li, and Wang]{song2024preference}
Song, F., Yu, B., Li, M., Yu, H., Huang, F., Li, Y., and Wang, H.
\newblock Preference ranking optimization for human alignment.
\newblock In \emph{Proceedings of the AAAI Conference on Artificial Intelligence}, volume~38, pp.\  18990--18998, 2024.

\bibitem[Stiennon et~al.(2020)Stiennon, Ouyang, Wu, Ziegler, Lowe, Voss, Radford, Amodei, and Christiano]{stiennon2020learning}
Stiennon, N., Ouyang, L., Wu, J., Ziegler, D., Lowe, R., Voss, C., Radford, A., Amodei, D., and Christiano, P.~F.
\newblock Learning to summarize with human feedback.
\newblock \emph{Advances in Neural Information Processing Systems}, 33:\penalty0 3008--3021, 2020.

\bibitem[Team et~al.(2023)Team, Anil, Borgeaud, Wu, Alayrac, Yu, Soricut, Schalkwyk, Dai, Hauth, et~al.]{team2023gemini}
Team, G., Anil, R., Borgeaud, S., Wu, Y., Alayrac, J.-B., Yu, J., Soricut, R., Schalkwyk, J., Dai, A.~M., Hauth, A., et~al.
\newblock Gemini: a family of highly capable multimodal models.
\newblock \emph{arXiv preprint arXiv:2312.11805}, 2023.

\bibitem[Team(2024)]{qwen1.5}
Team, Q.
\newblock Introducing qwen1.5, February 2024.
\newblock URL \url{https://qwenlm.github.io/blog/qwen1.5/}.

\bibitem[Wang et~al.(2024)Wang, Zheng, Chen, Liu, Dou, Huang, Shen, Jin, Zhou, Shi, et~al.]{wang2024secrets}
Wang, B., Zheng, R., Chen, L., Liu, Y., Dou, S., Huang, C., Shen, W., Jin, S., Zhou, E., Shi, C., et~al.
\newblock Secrets of rlhf in large language models part ii: Reward modeling.
\newblock \emph{arXiv preprint arXiv:2401.06080}, 2024.

\bibitem[Wei et~al.(2023{\natexlab{a}})Wei, Luan, Liu, Dong, and Wang]{wei2023cmath}
Wei, T., Luan, J., Liu, W., Dong, S., and Wang, B.
\newblock Cmath: Can your language model pass chinese elementary school math test?
\newblock \emph{arXiv preprint arXiv:2306.16636}, 2023{\natexlab{a}}.

\bibitem[Wei et~al.(2023{\natexlab{b}})Wei, Wang, Liu, Ding, and Zhang]{wei2023magicoder}
Wei, Y., Wang, Z., Liu, J., Ding, Y., and Zhang, L.
\newblock Magicoder: Source code is all you need.
\newblock \emph{arXiv preprint arXiv:2312.02120}, 2023{\natexlab{b}}.

\bibitem[Wu et~al.(2023)Wu, Zhu, Zhang, Wen, Ramchandran, and Jiao]{wu2023pairwise}
Wu, T., Zhu, B., Zhang, R., Wen, Z., Ramchandran, K., and Jiao, J.
\newblock Pairwise proximal policy optimization: Harnessing relative feedback for llm alignment.
\newblock \emph{arXiv preprint arXiv:2310.00212}, 2023.

\bibitem[Yang et~al.(2024)Yang, Yang, Hui, Zheng, Yu, Zhou, Li, Li, Liu, Huang, et~al.]{yang2024qwen2}
Yang, A., Yang, B., Hui, B., Zheng, B., Yu, B., Zhou, C., Li, C., Li, C., Liu, D., Huang, F., et~al.
\newblock Qwen2 technical report.
\newblock \emph{arXiv preprint arXiv:2407.10671}, 2024.

\bibitem[Yuan et~al.(2023)Yuan, Yuan, Tan, Wang, Huang, and Huang]{yuan2023rrhf}
Yuan, Z., Yuan, H., Tan, C., Wang, W., Huang, S., and Huang, F.
\newblock Rrhf: Rank responses to align language models with human feedback without tears.
\newblock \emph{arXiv preprint arXiv:2304.05302}, 2023.

\bibitem[Zhang et~al.(2024)Zhang, Chen, Chen, Shen, Sun, and Gan]{zhang2024improving}
Zhang, S., Chen, Z., Chen, S., Shen, Y., Sun, Z., and Gan, C.
\newblock Improving reinforcement learning from human feedback with efficient reward model ensemble.
\newblock \emph{arXiv preprint arXiv:2401.16635}, 2024.

\bibitem[Zhao et~al.(2020)Zhao, Shang, Liu, Wang, and Liu]{zhao2020ape210k}
Zhao, W., Shang, M., Liu, Y., Wang, L., and Liu, J.
\newblock Ape210k: A large-scale and template-rich dataset of math word problems.
\newblock \emph{arXiv preprint arXiv:2009.11506}, 2020.

\bibitem[Zhao et~al.(2023)Zhao, Joshi, Liu, Khalman, Saleh, and Liu]{zhao2023slic}
Zhao, Y., Joshi, R., Liu, T., Khalman, M., Saleh, M., and Liu, P.~J.
\newblock Slic-hf: Sequence likelihood calibration with human feedback.
\newblock \emph{arXiv preprint arXiv:2305.10425}, 2023.

\bibitem[Zhou et~al.(2024)Zhou, Jiang, Shen, Zhou, and He]{zhou2024leveraging}
Zhou, J., Jiang, C., Shen, W., Zhou, X., and He, X.
\newblock Leveraging web-crawled data for high-quality fine-tuning.
\newblock \emph{arXiv preprint arXiv:2408.08003}, 2024.

\bibitem[Zhu et~al.(2023)Zhu, Sharma, Frujeri, Dong, Zhu, Jordan, and Jiao]{zhu2023fine}
Zhu, B., Sharma, H., Frujeri, F.~V., Dong, S., Zhu, C., Jordan, M.~I., and Jiao, J.
\newblock Fine-tuning language models with advantage-induced policy alignment.
\newblock \emph{arXiv preprint arXiv:2306.02231}, 2023.

\bibitem[Zhu et~al.(2024)Zhu, Guo, Shao, Yang, Wang, Xu, Wu, Li, Gao, Ma, et~al.]{zhu2024deepseek}
Zhu, Q., Guo, D., Shao, Z., Yang, D., Wang, P., Xu, R., Wu, Y., Li, Y., Gao, H., Ma, S., et~al.
\newblock Deepseek-coder-v2: Breaking the barrier of closed-source models in code intelligence.
\newblock \emph{arXiv preprint arXiv:2406.11931}, 2024.

\bibitem[Ziegler et~al.(2019)Ziegler, Stiennon, Wu, Brown, Radford, Amodei, Christiano, and Irving]{ziegler2019fine}
Ziegler, D.~M., Stiennon, N., Wu, J., Brown, T.~B., Radford, A., Amodei, D., Christiano, P., and Irving, G.
\newblock Fine-tuning language models from human preferences.
\newblock \emph{arXiv preprint arXiv:1909.08593}, 2019.

\end{thebibliography}
\bibliographystyle{icml2025}

\newpage
\appendix
\onecolumn

\section{Reward Model}
\label{app:reward}

\begin{figure}[th]
     \centering
     \begin{subfigure}[b]{0.8\textwidth}
         \centering
         \includegraphics[width=\textwidth]{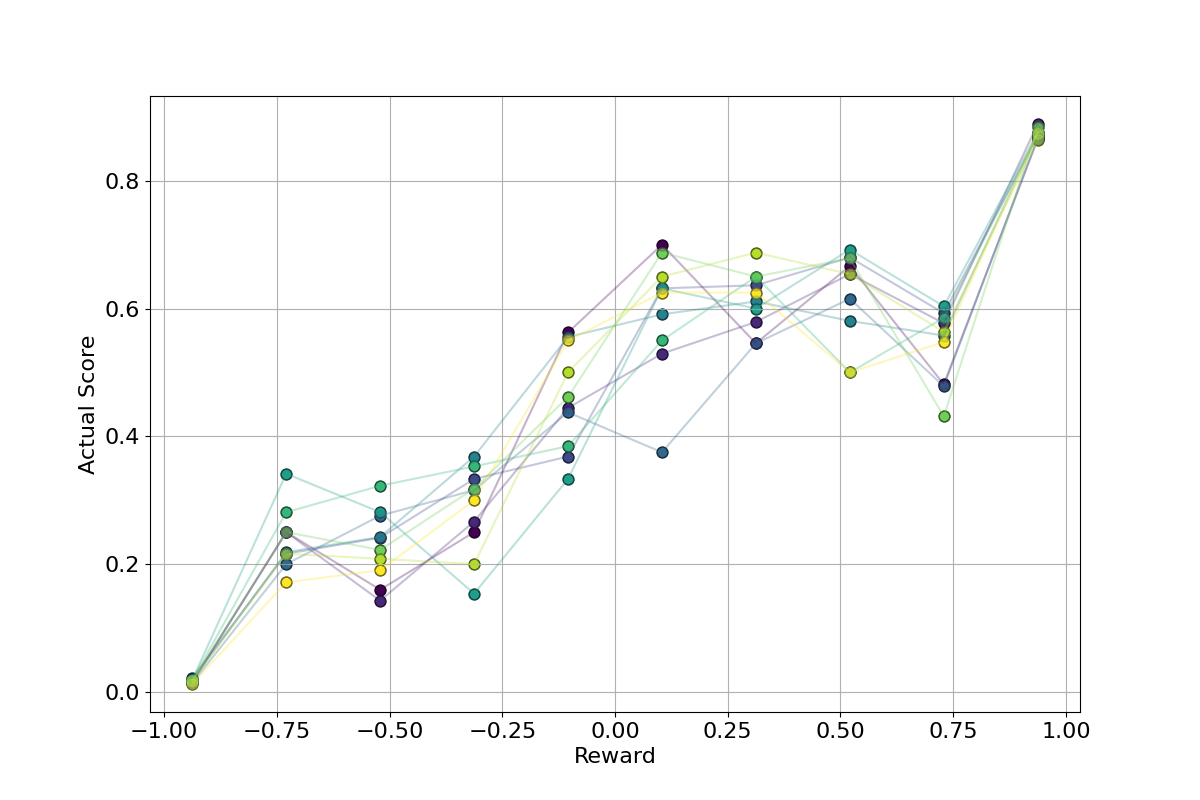}
         \caption{The actual scores vs. the reward values for the reward model evaluated on MBPP}
         \label{fig:reward_mbpp}
     \end{subfigure}
     \hfill
     \begin{subfigure}[b]{0.8\textwidth}
         \centering
         \includegraphics[width=\textwidth]{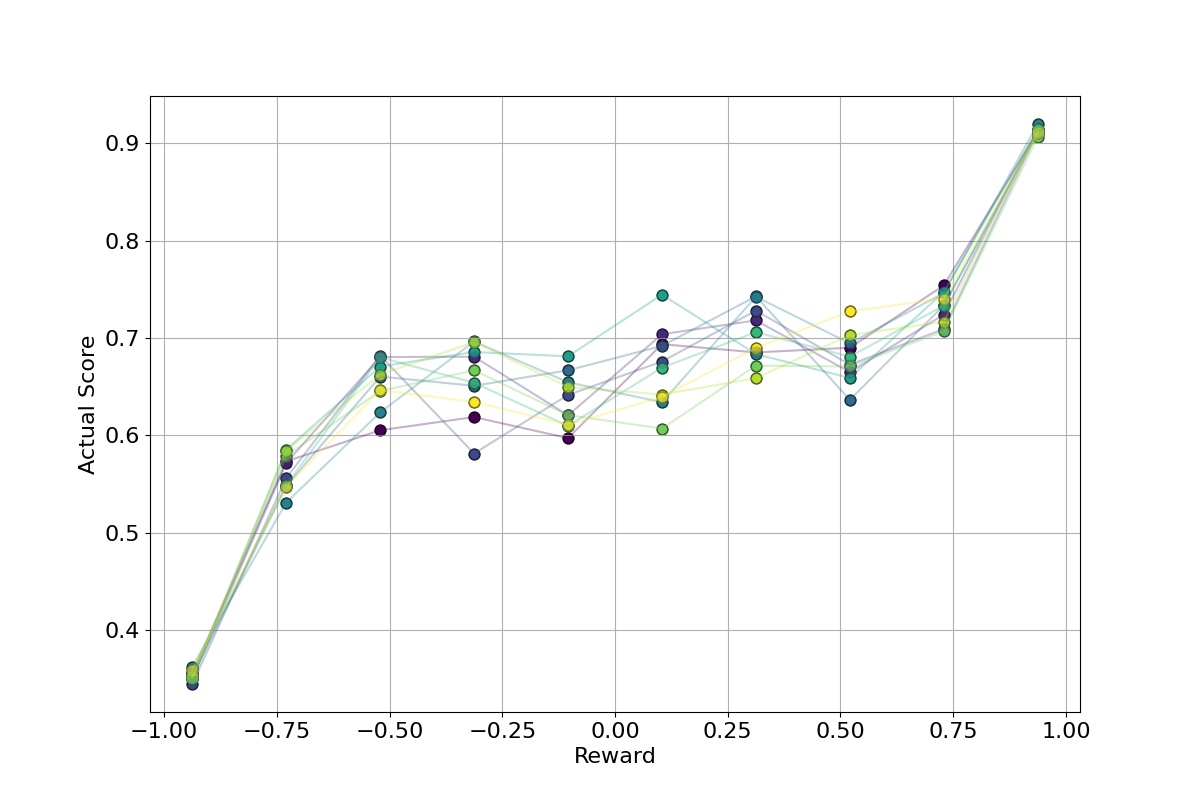}
         \caption{The actual scores vs. the reward values for the reward model evaluated on LeetCode}
         \label{fig:reward_leetcode}
     \end{subfigure}
        \caption{
    We provide additional evidence that the reward model is less reliable when it yields moderate rewards than when it yields high or low rewards.
    We conduct the same statistics as in Figure~\ref{fig:reward_model} but on different benchmarks.
    Specifically, 
    the reward models for the MBPP and LeetCode benchmarks are trained separately using the corresponding datasets for these two benchmarks.
    The MBPP and LeetCode benchmarks contains 378 and 1570 prompts respectively and we collect 10 responses for each prompt using a fine-tuned policy.
    We group the responses with similar rewards and calculate the average of their actual scores (i.e., the average correctness), indicating each group by one point.
    To evaluate the reliability of the reward model, we repeat the process ten times resulting in ten lines.}
    \label{fig:app_reward}
\end{figure}

\begin{figure}[th]
     \centering
     \begin{subfigure}[b]{0.8\textwidth}
         \centering
         \includegraphics[width=\textwidth]{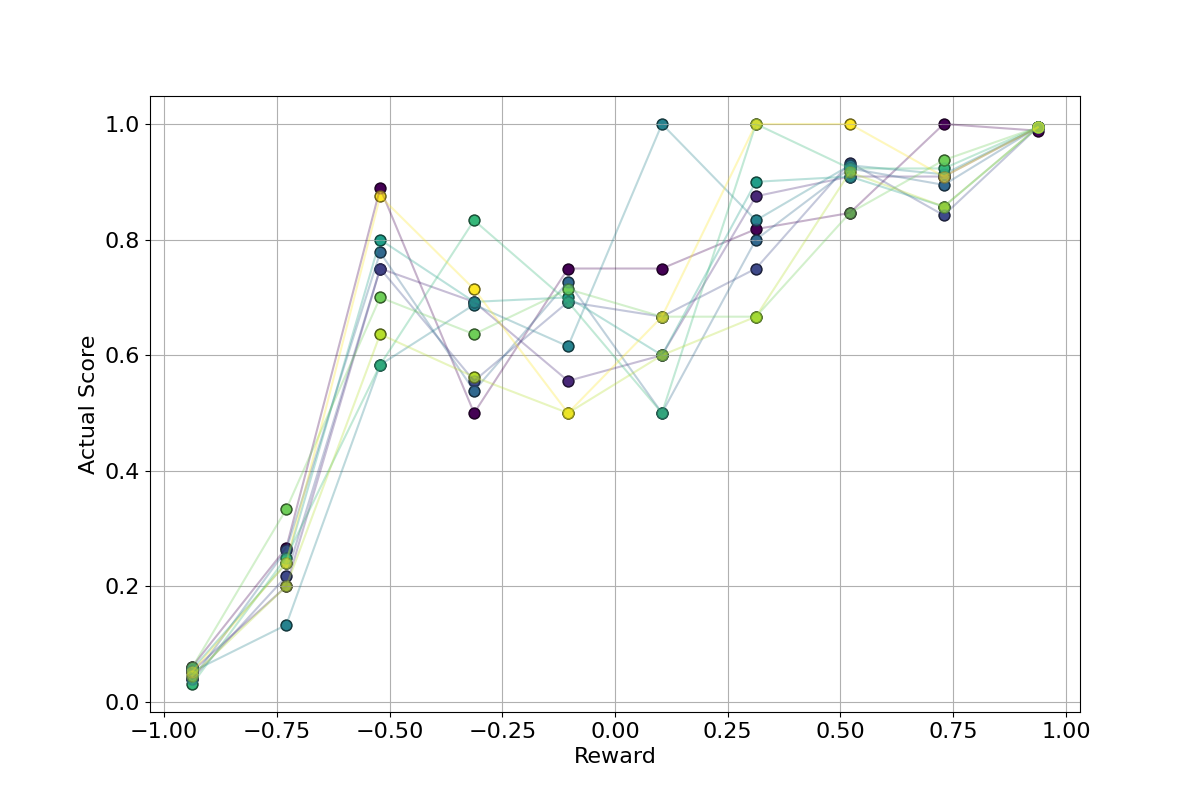}
         \caption{The actual scores vs. the reward values for the reward model evaluated on Ape210k}
         \label{fig:reward_ape201k}
     \end{subfigure}
     \hfill
     \begin{subfigure}[b]{0.8\textwidth}
         \centering
         \includegraphics[width=\textwidth]{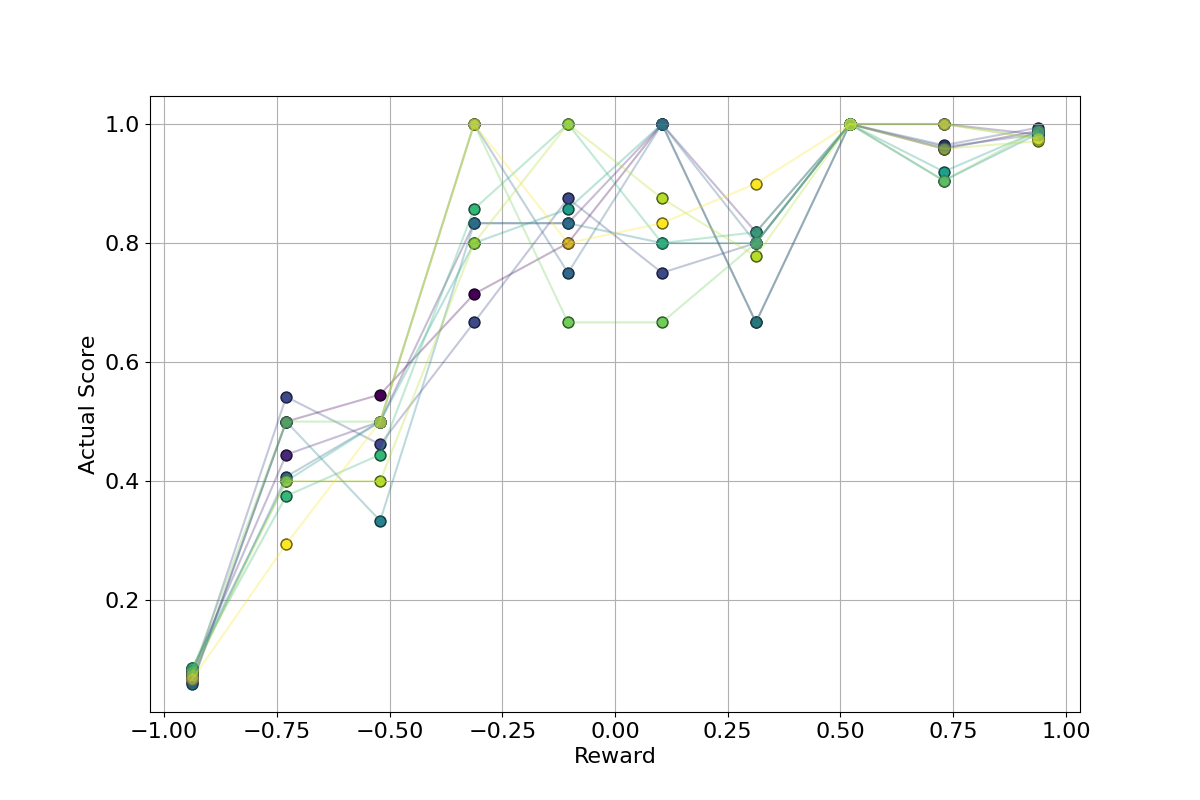}
         \caption{The actual scores vs. the reward values for the reward model evaluated on CMATH}
         \label{fig:reward_cmath}
     \end{subfigure}
        \caption{
    We provide additional evidence that the reward model is less reliable when it yields moderate rewards than when it yields high or low rewards.
    We conduct the same statistics as in Figure~\ref{fig:reward_model} but on different benchmarks.
    Specifically, 
    the reward models for the Ape210k and CMATH benchmarks are trained separately using the corresponding datasets for these two benchmarks.
    We collect 10 responses for each prompt in the dataset using a fine-tuned policy.
    We group the responses with similar rewards and calculate the average of their actual scores (i.e., the average correctness), indicating each group by one point.
    To evaluate the reliability of the reward model, we repeat the process ten times resulting in ten lines.}
    \label{fig:app_reward_math}
\end{figure}

The design of our algorithm is motivated by the observation that the reward model is less reliable when it yields moderate rewards. 
To provide more evidence that this property is universal across a broader range of benchmarks,
we also analyze the reward function on different benchmarks of code generation (MBPP and LeetCode) and math reasoning (Ape210K~\citep{zhao2020ape210k} and CMATH~\citep{wei2023cmath}).
We repeat the process in Figure~\ref{fig:reward_model} on these benchmarks and plot the figures in Figure~\ref{fig:app_reward} and Figure~\ref{fig:app_reward_math}.
Note that we train different reward functions based on the datasets from these two benchmarks.
We observe that the property holds on these four additional benchmarks across different tasks, indicating this property may extend to broader fields.

Intuitively, this property should be universal to a broader range of tasks, e.g., on Helpfulness and Harmlessness tasks \citep{bai2022training}.
For code generation tasks, it is quite common that some samples (e.g., the response matches the known correct answer or the response contains an obvious error) are easier to evaluate than others (e.g., the response tries to solve the problem by a novel approach). 
Therefore, those samples that are hard to evaluate by human should also be hard instances for the reward model.

\clearpage
\section{Qualitative results}
\label{app:case}

In this section, we provide qualitative results on 1) how responses with high/middle/low rewards look like and why responses with middle rewards are unreliable; and 2) the qualitative difference between the code generated by the PF-PPO policy and the standard PPO (PPO-S) policy.

\subsection{Analysis on the the responses associated with different rewards}

We present a prompt along with several responses, including a correct response but assigned with a low reward, an incorrect response but assigned with a high reward, an incorrect response with a low reward, and a correct response with a high reward.
The prompt describes a coding problem that requires to convert fractions to decimals.

We have the following findings: 
\begin{itemize}  
    \item For the correct response but assigned with a low reward, the generated code is less clear and harder to read. For example, the code may mix several steps into one line. 
    \item For the incorrect response but assigned with a high reward, the response incorrectly mixes two correct approaches. This mistake can hardly be identified by the reward model (and even GPT-4).
    \item For the incorrect response assigned with a low reward, the response contains an obvious mistake which is easily detected by the reward model.
\end{itemize}

We also provide detailed analysis into the solutions to this problem.
The given prompt is a coding task to convert fraction to decimal.
\begin{lstlisting}[breaklines,numbers=left]
Given two integers representing the numerator and denominator of a fraction, return the fraction in string format.
If the fractional part is repeating, enclose the repeating part in parentheses.
If multiple answers are possible, return any of them.
It is guaranteed that the length of the answer string is less than 104 for all the given inputs.

Please complete the following python code precisely:
```python
class Solution:
    def fractionToDecimal(self, numerator: int, denominator: int) -> str:
```
\end{lstlisting}

This is a correct response with a high reward.
This solution is thorough with clear variable names and properly structured steps.
It is easier to read due to breaking down steps explicitly such as calculating the integer part and handling the remainder.

\begin{lstlisting}[breaklines,numbers=left]


To solve the problem of converting a fraction to its decimal representation, we can follow these steps:

1. **Determine the sign of the resultant decimal**: If both the numerator and the denominator are negative, the result should be positive. Otherwise, the result will have a leading minus sign if and only if either the numerator or the denominator is negative.

2. **Convert the numerator and denominator to positive values**: Since the division process involves integer division and the decimal part requires modulus operation, both numerator and denominator should be made positive for easier computation.

3. **Calculate the integer part**: The integer part is the result of integer division of the numerator by the denominator.

4. **Manage the remainder**: As we perform integer division, the remainder will be used to form the fractional part. This remainder will be treated as the new numerator for the next step.

5. **Use a dictionary to manage repetition**: To detect recurring decimals, we need to keep track of all remainders we have seen and their respective positions in the decimal expansion. If a remainder repeats, it indicates a recurring decimal.

6. **Build the result string**: Start with the integer part, then append a decimal point. Then, use a loop to compute each digit of the fractional part. Use the dictionary to detect and mark any recurring decimals.

7. **Return the result**: Finally, return the constructed string for the result.

Below is the complete Python code for the `fractionToDecimal` method:

```python
class Solution:
    def fractionToDecimal(self, numerator: int, denominator: int) -> str:
        if numerator == 0:
            return "0"
        
        # Determine the sign of the result
        result_sign = "-" if (numerator < 0) ^ (denominator < 0) else ""
        
        # Make both numerator and denominator positive
        numerator = abs(numerator)
        denominator = abs(denominator)
        
        # Calculate the integer part of the result
        integer_part = numerator // denominator
        remainder = numerator % denominator
        
        # If there's no remainder, return the integer part
        if remainder == 0:
            return result_sign + str(integer_part)
        
        # Prepare to build the fractional part
        result = []
        result.append(".")  # Start with the decimal point
        remainder_map = {}  # To keep track of remainders and their positions
        
        # Process each place value in the fractional part
        position = 0
        while remainder not in remainder_map:
            remainder_map[remainder] = position
            remainder *= 10
            result_digit = remainder // denominator
            result.append(str(result_digit))
            remainder %= denominator
            if remainder == 0:
                break
            position += 1
        
        # If there's a repeat, insert parentheses
        if remainder in remainder_map:
            repeat_start = remainder_map[remainder]
            result.insert(repeat_start + 1, "(")
            result.append(")")
        
        return result_sign + str(integer_part) + "".join(result)
```

This code correctly handles edge cases like zero denominator, negative numbers, repeating decimals, and ensures that the result is less than 10^4 characters long.
\end{lstlisting}

This is a correct response but assigned with a low reward.
Compared with the previous response with high reward, this response mixes multiple operations in one line, making it harder to understand (e.g., Line 34).

\begin{lstlisting}[breaklines,numbers=left]
To solve the problem of converting a fraction to its decimal representation, we can use the following approach:

1. Handle the sign of the result.
2. Calculate the integral part of the result, which is the part before the decimal point.
3. Handle the case where the fractional part is repeating.
4. Calculate the fractional part and handle cycles.

Here's the completed Python code:

```python
class Solution:
    def fractionToDecimal(self, numerator: int, denominator: int) -> str:
        # Handle sign of result
        sign = '-' if numerator * denominator < 0 else ''
        numerator, denominator = abs(numerator), abs(denominator)
        
        # Calculate integral part
        integral_part = numerator // denominator
        
        # Calculate fractional part
        remainder = numerator % denominator
        if remainder == 0:
            return sign + str(integral_part)
        
        # Use a dictionary to track remainders and their positions
        # in the fractional part to detect cycles
        seen_remainders = {}
        result = []
        index = 0
        
        while remainder not in seen_remainders:
            seen_remainders[remainder] = index
            result.append(str(remainder * 10 // denominator))
            remainder = (remainder * 10) % denominator
            if remainder == 0:
                break
            index += 1
        
        # If there's a cycle, insert the parentheses
        if remainder in seen_remainders:
            index = seen_remainders[remainder]
            result.insert(index, '(')
            result.append(')')
        
        return sign + str(integral_part) + '.' + ''.join(result)
```

### Explanation:

- We first handle the signs of the numerator and denominator.
- We then calculate the integral part of the result by integer division.
- If the remainder is zero, the fraction is an integer and we return it as a string.
- We use a dictionary `seen_remainders` to keep track of the positions where each remainder appears, allowing us to detect cycles.
- If a cycle is detected, we insert parentheses at the start of the repeating part.
- Finally, we assemble the result as a string and return it.
\end{lstlisting}

This is an incorrect response but assigned with a high reward.
In Line 32, the decimal point is added to the \texttt{result} list but is not later counted when getting wrapped by the parentheses, leading to the wrong format.
This is a mixture of two correct approaches, one that adds the decimal points to \texttt{result} but sets an offset for this (cf. Line 44 and Line 62 in the first response) and one that outputs the decimal point separately (cf. Line 45 in the second response).
\begin{lstlisting}[breaklines,numbers=left]
To solve the problem of converting a fraction to a decimal string, we need to handle the following cases:

1. **Integers**: If the numerator and denominator are both integers, the decimal representation is straightforward: divide the numerator by the denominator and convert the result to a string.

2. **Decimal Representation**: If the division produces a remainder, it indicates a repeating decimal pattern. Use a dictionary to keep track of the remainders and their corresponding positions in the resultant decimal string. Whenever a remainder repeats, enclose the repeating part in parentheses.

3. **Handling Negative Numbers**: Handle both positive and negative numbers appropriately.

Here is the complete code for the solution:

```python
class Solution:
    def fractionToDecimal(self, numerator: int, denominator: int) -> str:
        # Handle sign of the result
        sign = '-' if numerator * denominator < 0 else ''
        numerator, denominator = abs(numerator), abs(denominator)
        
        # Calculate integral part
        integral_part = numerator // denominator
        numerator %= denominator
        
        # Handle zero remainder
        if numerator == 0:
            return sign + str(integral_part)
        
        # Prepare list to store the decimal digits
        result = []
        # Dictionary to keep track of remainders and their positions
        remainder_map = {}
        
        # Initialize the decimal part
        result.append('.')
        position = 0
        
        # Loop until the remainder repeats or becomes zero
        while numerator > 0:
            # If the remainder has already appeared before, it's a repeating part
            if numerator in remainder_map:
                result.insert(remainder_map[numerator], '(')
                result.append(')')
                break
            
            # Record the position of the current remainder
            remainder_map[numerator] = position
            numerator *= 10
            result.append(str(numerator // denominator))
            numerator %= denominator
            position += 1
        
        return sign + str(integral_part) + ''.join(result)
```

### Explanation:
- We use a dictionary `remainder_map` to track the position of each remainder seen. This helps us identify the repeating part of the decimal if it exists.
- The loop continues until the remainder becomes zero or repeats itself.
- If the remainder repeats, we insert the opening `(` immediately before the repeated part, and close it with `)`.
- The result is assembled from the integral part, the decimal point, and the decimal digits or parenthesized pattern.
\end{lstlisting}

This is an incorrect response with a low reward.
In Line 59-61, the program contains an obvious error that it cannot handle the scenario where there exists a decimal part but does not contain any repeating part.

\begin{lstlisting}[breaklines,numbers=left]


To solve the problem of converting a fraction to its decimal representation, we need to consider the following steps:

1. **Handling Zero Denominator**: If the denominator is zero, it's impossible to divide, so we should return "NaN" (Not a Number) or handle it according to the problem constraints.

2. **Handling Negative Numbers**: If either the numerator or denominator is negative, the sign of the result should be negative.

3. **Calculate the Integral Part**: The integral part is the part before the decimal point, which can be calculated by integer division.

4. **Calculate the Fractional Part**:
   - Use a dictionary to keep track of the positions where the same remainder started repeating.
   - Divide the remainder by the denominator repeatedly, building the fractional part digit by digit.
   - If the remainder repeats, it indicates a recurring decimal, and we should insert parentheses to mark it.

Here's the complete code implementation:

```python
class Solution:
    def fractionToDecimal(self, numerator: int, denominator: int) -> str:
        # Handle zero denominator
        if denominator == 0:
            return "NaN"
        
        # Handle signs
        result_sign = '-' if (numerator < 0) ^ (denominator < 0) else ''
        numerator, denominator = abs(numerator), abs(denominator)
        
        # Calculate the integral part
        integral_part = numerator // denominator
        numerator %= denominator
        
        # Initialize the fractional part list and a dictionary to track remainders
        fractional_part = []
        remainder_index = {}
        
        # Continue while there is a remainder
        position = 0
        while numerator > 0:
            # If the remainder has already occurred before, it's a repeating decimal
            if numerator in remainder_index:
                break
            
            # Record this remainder's position
            remainder_index[numerator] = position
            
            # Calculate the next digit
            numerator *= 10
            fractional_part.append(str(numerator // denominator))
            numerator %= denominator
            position += 1
        
        # Build the result
        if not fractional_part:
            # No repeating part, return integral part
            return result_sign + str(integral_part)
        else:
            # Repeating part detected
            repeating_start = remainder_index[numerator]
            return result_sign + str(integral_part) + '.' + ''.join(fractional_part[:repeating_start]) + \
                   '(' + ''.join(fractional_part[repeating_start:]) + ')'
```

### Explanation:

- **Integral Part**: We calculate the integral part as `numerator // denominator`.
- **Non-repeating Part**: We use a dictionary to track positions of remainders, and if the same remainder appears again, we know the fraction is repeating.
- **Repeating Part**: When a repeating part is detected, we insert parentheses around the repeating part.
- **Edge Cases**: We handle edge cases such as negative numbers, zero denominator, and ensuring the result is within the constraints of the problem.

This implementation correctly handles both whole number and repeating decimal results, adhering to the problem statement's requirements.
\end{lstlisting}

\subsection{Analysis on the responses generated by PF-PPO compared with standard PPO (PPO-S)}

We compare the answers from PF-PPO and PPO-S respectively for the same prompts, and conclude that the answer from the PF-PPO policy is more inclined to follow a standard approach and the response is more concise (or shorter), making it easier to understand and implement.


Specifically, we present their responses for the prompt that request the agent to write a code to \emph{find all safe nodes in a directed graph}.
The two responses given by PF-PPO and PPO-S are both correct. 
However, the difference is that PF-PPO adopts the deep first search (DFS) method while PPO-S adopts the topological sorting approach.
The logic is simpler for DFS and the implementation is easier, making it easier to understand.
Both approaches have roughly equivalent time and space costs, but the DFS method is slightly more space efficient.


These findings suggest an advantage in terms of readability and implementation simplicity when using the PF-PPO policy (e.g., using recursion instead of using a queue to track BFS).

Prompt (find all safe nodes in a directed graph):
\begin{lstlisting}[breaklines,numbers=left]
There is a directed graph of n nodes with each node labeled from 0 to n - 1. The graph is represented by a 0-indexed 2D integer array graph where graph[i] is an integer array of nodes adjacent to node i, meaning there is an edge from node i to each node in graph[i].
A node is a terminal node if there are no outgoing edges. A node is a safe node if every possible path starting from that node leads to a terminal node (or another safe node).
Return an array containing all the safe nodes of the graph. The answer should be sorted in ascending order.

Please complete the following python code precisely:
```python
class Solution:
    def eventualSafeNodes(self, graph: List[List[int]]) -> List[int]:
```
\end{lstlisting}

The response from PF-PPO:
\begin{lstlisting}[breaklines,numbers=left]


To solve the problem of finding all safe nodes in a directed graph, we can use a topological sorting approach. The idea is to explore the graph using depth-first search (DFS) and maintain a status for each node indicating whether it is safe or not. Here's the step-by-step breakdown of the solution:

1. **Initialize Status**: We will use a status array to track the safety of each node. 0 means the node is not visited yet, 1 means it is a safe node, and -1 means it is currently in the recursion stack (i.e., being processed in the current DFS path).

2. **DFS Function**:
   - For each node, if it's safe or not is already determined (i.e., status is not 0), return the status.
   - Set the status to -1 to indicate that we are currently processing it.
   - For each neighbor of the current node, call DFS. If any neighbor returns false, it indicates that this path has a cycle and the node is not safe.
   - If no cycle is found for all neighbors, mark the node as safe and return true.

3. **Iterate over all nodes**: Starting from each node that is not visited, perform a DFS.

4. **Collect Results**: All nodes that return true from DFS are safe nodes. Collect these node indices and return them in ascending order.

Here's the complete Python code for the solution:

```python
from typing import List

class Solution:
    def eventualSafeNodes(self, graph: List[List[int]]) -> List[int]:
        n = len(graph)
        # status: 0 for unvisited, 1 for safe, -1 for processing
        status = [0] * n
        
        def dfs(node):
            # If the node is already processed (i.e., marked as safe or unsafe), return its status
            if status[node]!= 0:
                return status[node] == 1
            
            # Mark the node as being processed
            status[node] = -1
            
            # Visit all neighbors
            for neighbor in graph[node]:
                # If any neighbor is unsafe or in a cycle, the node is unsafe
                if not dfs(neighbor):
                    return False
            
            # All neighbors are safe, mark the node as safe and return true
            status[node] = 1
            return True
        
        # Try to DFS from each node
        safe_nodes = []
        for i in range(n):
            if dfs(i):
                safe_nodes.append(i)
        
        return safe_nodes
```

### Explanation:
- **Topological Sorting**: We are using DFS to check if a path exists from a node to a terminal node. If a cycle is detected, the node is not safe.
- **DFS Function**: The `dfs` function determines if a node is safe by checking all of its neighbors recursively. If any neighbor is unsafe or in a cycle, the node is unsafe.
- **Status Tracking**: The status array is used to track if a node is safe or not, and helps avoid redundant computations.
- **Result Collection**: All nodes that return `True` from the `dfs` function are safe nodes, and are added to the result list.

This approach efficiently identifies safe nodes by using a topological sorting approach and a depth-first search approach to detect cycles.
\end{lstlisting}

The response from PPO-S:
\begin{lstlisting}[breaklines,numbers=left]


To solve this problem, we can use a \red{topological sort approach} combined with a depth-first search (DFS) approach to determine the terminal and safe nodes in the graph. The topological sort ensures that we process nodes in the order of their out-degree, which is suitable for directed acyclic graphs (DAGs), where no cycles can form.

The idea is to traverse the graph using DFS, keeping track of the current path during traversal to detect cycles. Nodes that do not lead to a cycle are considered terminal, and nodes that are part of a cycle or are part of a path that leads to a cycle are considered unsafe.

Here's the Python implementation of the solution:

```python
from typing import List
from collections import defaultdict, deque

class Solution:
    def eventualSafeNodes(self, graph: List[List[int]]) -> List[int]:
        # Initialize a dictionary to store the out-degree of each node
        out_degree = defaultdict(int)
        # Initialize a dictionary to store the dependencies (incoming edges) of each node
        dependencies = defaultdict(list)
        
        n = len(graph)
        # Build the dependencies and out-degree dictionaries
        for u in range(n):
            out_degree[u] = len(graph[u])
            for v in graph[u]:
                dependencies[v].append(u)
        
        # Initialize a queue for BFS and a set to keep track of safe nodes
        queue = deque()
        safe_nodes = set()
        
        # Add all nodes with no outgoing edges (terminals) to the queue
        for u in range(n):
            if out_degree[u] == 0:
                queue.append(u)
        
        # BFS to find safe nodes
        while queue:
            u = queue.popleft()
            safe_nodes.add(u)
            # For each node v that depends on u, decrement its out-degree
            # If v's out-degree becomes zero, it becomes a terminal
            for v in dependencies[u]:
                out_degree[v] -= 1
                if out_degree[v] == 0:
                    queue.append(v)
        
        # Sort and return safe nodes
        return sorted(safe_nodes)
```

### Explanation:
1. **Dependencies**: We build a `dependencies` dictionary, where each key is a node and the values are a list of nodes that depend on the key. This is used to track back edges.
2. **Out-degree**: We also keep track of the out-degree of each node. This is used to determine which nodes are terminal (i.e., have no outgoing edges).
3. **Topological Sort (BFS)**: We start by adding all nodes with no outgoing edges (terminals) to the queue. We then process each node from this queue. For each node, we mark it as safe, reduce the out-degree of its dependencies, and if any of those dependencies become terminals (i.e., have no outgoing edges), they are added to the queue. This ensures we process all nodes that are part of cycles and prevent future revisits to nodes involved in such cycles.
4. **Result**: The safe nodes are collected and sorted and returned as the final result.

This approach ensures that we correctly identify terminal nodes and use them to establish safe nodes, as well as handle cycles in the graph.
\end{lstlisting}

\section{Discussion on the computational costs}
\label{app:computation_cost}

PPO-S, PPO-M, and PF-PPO all collect different number of responses per query and train using different number of samples.
For clarity, we list the computational complexity of these algorithms in Table~\ref{tab:sampleeff}.
Note that, for all algorithms, we select the best checkpoint on the evaluation set and report the performance of this checkpoint.
Combining the results in Table~\ref{tab:sampleeff} and Figure~\ref{fig:details}, we can draw the following conclusions:
First, the total computational complexity of PPO-S and PPO-M is almost the same, and the only difference is that PPO-M is more likely to learn from different responses with the same query in the same batch or adjacent batches, which improves the performance. 
Second, the computational complexity of PF-PPO is less than that of PPO-S and PPO-M, while PF-PPO outperforms them.
This indicates the effectiveness of our method.

\begin{table*}[b]
\centering
\begin{tabular}{lrrr}
\toprule[1pt]
\textbf{} & \textbf{PPO-S} & \textbf{PPO-M} & \textbf{PF-PPO (BR / BW)} \\
\midrule
Queries sampled per iteration & $5n$ & $n$ & $n$ \\
Responses sampled per query & $1$ & $5$ & $5$ \\
\#Query-response pairs per iteration & $5n$ & $5n$ & $5n$ \\
Reward model forward pass per iteration & $5n$ & $5n$ & $5n$ \\
Critic forward\&backward pass per iteration & $5nm$ & $5nm$ & $2nm$ \\
Policy forward\&backward pass per iteration & $5nm$ & $5nm$ & $2nm$ \\
\midrule
HumanEval   & 100\% & +2.69\% & +6.15\% / +5.51\% \\
MBPP        & 100\% & +1.63\% & +2.85\% / +3.25\% \\
LeetCode    & 100\% & +18.25\% & +30.95\% / +20.63\%\\
\bottomrule[1pt]
\end{tabular}
\caption{Comparison of computational complexity and the performance of PPO-S, PPO-M, and PF-PPO. 
We use $n$ to denote the number of queries in the PPO query dataset, and use $m$ to denote the number of PPO epochs (i.e., each query-response pair is used to accumulate loss and gradient for $m$ times on average).
PPO-M and PF-PPO collect $N=5$ responses per query, and PF-PPO select $2$ out of the $N=5$ responses (on average) for network update.
We also show the performance improvement of PPO-M and PF-PPO based on PPO-S.
}
\label{tab:sampleeff}
\end{table*}




\end{document}